\documentclass[10pt,twocolumn,letterpaper]{article}

\usepackage{cvpr}              

\usepackage[dvipsnames]{xcolor}

\definecolor{cvprblue}{rgb}{0.21,0.49,0.74}
\usepackage[pagebackref,breaklinks,colorlinks,citecolor=cvprblue]{hyperref}
\usepackage[utf8]{inputenc} 
\usepackage[T1]{fontenc}    
\usepackage{hyperref}       
\usepackage{url}            
\usepackage{booktabs}       
\usepackage{amsfonts}       
\usepackage{nicefrac}       
\usepackage{microtype}      
\usepackage{xcolor}         
\usepackage{graphicx}
\usepackage{amssymb}
\usepackage{amsmath}
\usepackage{multirow}
\usepackage{multicol}
\usepackage{booktabs}
\usepackage{balance}
\usepackage{adjustbox}
\usepackage{caption}
\usepackage{subcaption}
\usepackage[font=small]{caption} 
\usepackage{soul}
\usepackage{bm}
\usepackage{times}
\usepackage{epsfig}
\usepackage{mathrsfs}
\usepackage{array}
\usepackage{makecell}

\def\paperID{9543} 
\def\confName{CVPR}
\def\confYear{2024}

\title{Towards Accurate Post-training Quantization for Diffusion Models}

\author{Changyuan Wang\textsuperscript{1}, ~Ziwei Wang\textsuperscript{3}, ~Xiuwei Xu\textsuperscript{2}, ~Yansong Tang\textsuperscript{1}\thanks{Corresponding author.}, ~Jie Zhou\textsuperscript{2}, ~Jiwen Lu\textsuperscript{2}\\
\textsuperscript{1}Shenzhen Key Laboratory of Ubiquitous Data Enabling, Shenzhen International Graduate School, \\Tsinghua University, China 
~\textsuperscript{2}Department of Automation, Tsinghua University, China \\
~\textsuperscript{3}Carnegie Mellon University \\
{\tt\small \{wangchan22@mails.,xxw21@mails.,tang.yansong@sz.,jzhou@,lujiwen@\}tsinghua.edu.cn;} \\
{\tt\small ziweiwa2@andrew.cmu.edu} \\
}

\begin{document}
\maketitle
\begin{abstract}
  In this paper, we propose an accurate post-training quantization framework of diffusion models (APQ-DM) for efficient image generation. Conventional quantization frameworks learn shared quantization functions for tensor discretization regardless of the generation timesteps in diffusion models, while the activation distribution differs significantly across various timesteps. Meanwhile, the calibration images are acquired in random timesteps which fail to provide sufficient information for generalizable quantization function learning. Both issues cause sizable quantization errors with obvious image generation performance degradation. On the contrary, we design distribution-aware quantization functions for activation discretization in different timesteps and search the optimal timesteps for informative calibration image generation, so that our quantized diffusion model can reduce the discretization errors with negligible computational overhead. Specifically, we partition various timestep quantization functions into different groups according to the importance weights, which are optimized by differentiable search algorithms. We also extend structural risk minimization principle for informative calibration image generation to enhance the generalization ability in the deployment of quantized diffusion model. Extensive experimental results show that our method outperforms the state-of-the-art post-training quantization of diffusion model by a sizable margin with similar computational cost\footnote{Code is available at https://github.com/ChangyuanWang17/APQ-DM}. 
\end{abstract}

\section{Introduction}
\label{sec:intro}

\begin{figure}[t]
	\centering
	\includegraphics[width=0.94\linewidth]{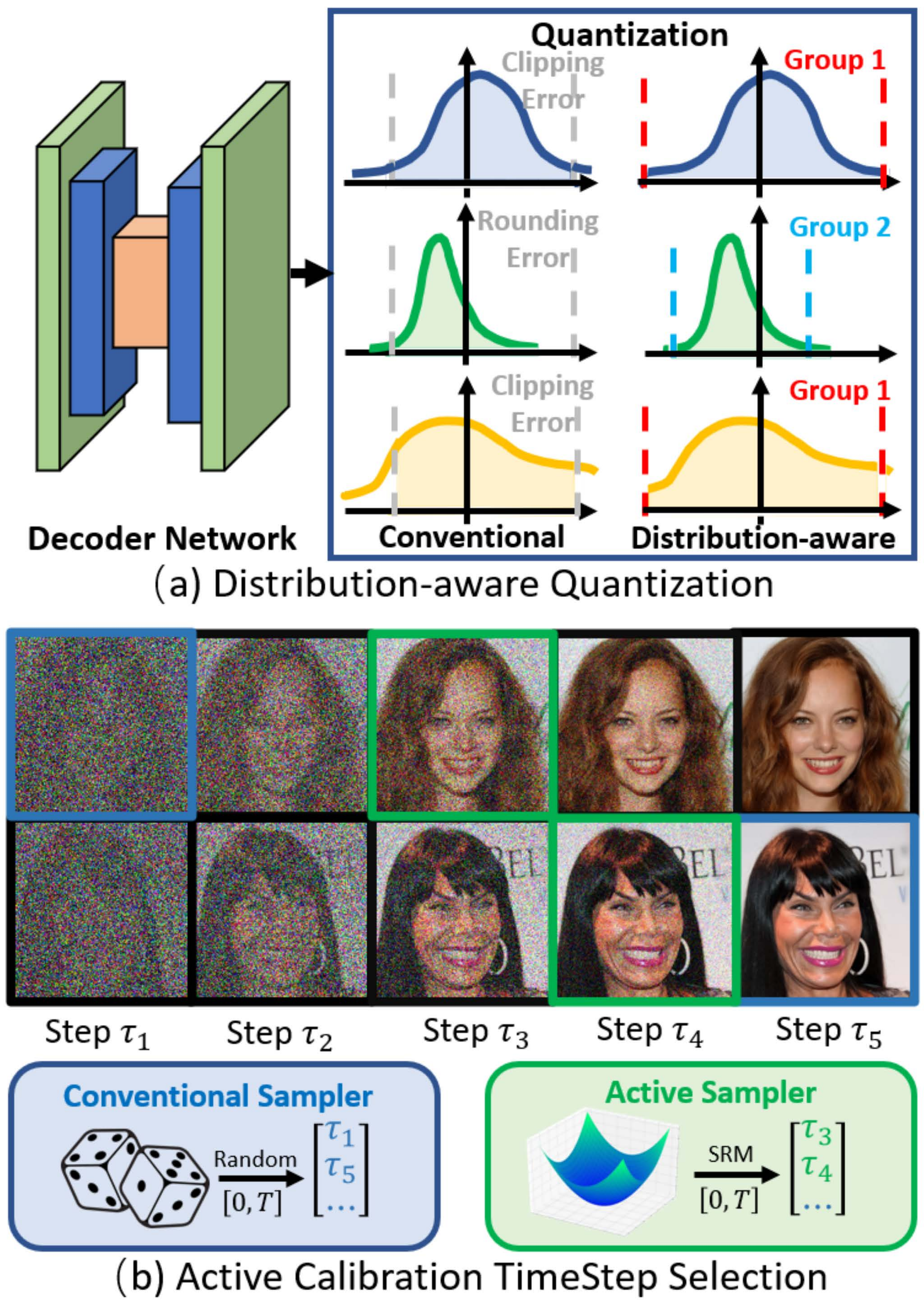}
        \vspace{-0.25cm}
	\caption{(a) Existing methods leverage shared quantization for activation discretization across different timesteps with significant quantization errors, while we divide timesteps into groups with specific rounding functions for each partition. (b) Conventional methods construct calibration set by randomly image selecting with ineffective supervision, while we actively sample timesteps based on the structural risk minimization (SRM) principle.}
	\vspace{-0.35cm}    
	\label{comparison}
\end{figure}

Denoising diffusion generative models \cite{ho2020denoising,song2020denoising} have achieved outstanding performance in a wide variety of computer vision tasks such as image edition \cite{avrahami2022blended,nichol2021glide}, style transformation \cite{su2022dual,yang2023zero}, image super-resolution \cite{saharia2022image,li2022srdiff} and many others. Compared with the generative adversarial networks (GAN), diffusion models obtained recovered contents with better quality and diversity on most downstream tasks. However, diffusion models usually require hundreds of noise evaluation steps to generate high-quality images from Gaussian noises by neural networks with millions of parameters, and the numerous forward passes in network inference result in heavy computation burden. Therefore, designing lightweight denoising process for diffusion models is highly demanded for flexible deployment in practical applications with limited resources such as mobile phones and robots.

To accelerate the generation process of diffusion models, recent studies made significant efforts in decreasing the sampling times of image denoising process \cite{bao2022analytic,song2020score,jolicoeur2021gotta} and reducing the network complexity in noise estimation of each step \cite{shang2022post,li2023q,li2022efficient}. The former removes redundant steps in the reverse process of diffusion, and the latter extends the network compression techniques to noise estimation such as pruning \cite{zhao2019variational,hu2016network} and quantization \cite{lee2021network,hubara2016binarized} for acceleration. We focus on the latter by quantizing the noise estimation networks with integer arithmetic inference. Due to the intractability of the training data and the unbearable training cost of diffusion models to fully optimize quantized network parameters, the post-training quantization framework for the pre-trained full-precision decoders is leveraged that only learns the rounding function parameters. Nevertheless, conventional data-free post-training quantization methods \cite{cai2020zeroq,zhong2022intraq} learn a shared layer-wise rounding function for all generation timesteps where the activation distribution varies obviously in diffusion models, and the calibration images are generated in random timesteps which fails to provide sufficient information to acquire generalizable quantization function. Consequently, both the inaccurate quantization functions and uninformative calibration images lead to significant quantization errors in noise estimation process, which degrades the synthesis performance by a sizable margin.

In this paper, we present an accurate post-training quantization framework for diffusion models in order to achieve efficient image generation. Different from existing methods that leverage shared layer-wise quantization functions for all timesteps and synthesizing calibration images in random timesteps for training, we partition timesteps into different groups to impose specific rounding functions for each group and sampling the optimal timesteps to generate informative calibrate images for quantization parameter learning. Therefore, the significant quantization errors of noise estimation in diffusion model deployment can be reduced with only negligible computation overhead. More specifically, we employ a differentiable search strategy to acquire the optimal group assignment for different generation timesteps, and learns individual rounding functions for each group with minimized discretization errors. For the differentiable search, the activations quantized by discretization functions in different groups are summed with learnable importance weights. We also generalize the structural risk minimization (SRM) principle for timestep selection to generate informative calibration images, where the entropy of rounding function weights in differentiable search and the sampling times of the timestep are considered as the criteria based on our formulation. Figure \ref{comparison} demonstrates the comparison between our method and conventional data-free post-training quantization framework for diffusion models. Extensive experimental results on unconditioned synthesis and conditional image generation across various network architectures clearly demonstrate that our method sizably increases the quality of the generated images with only negligible computational complexity. Our contributions can be summarized as follows:
\begin{itemize}
    \item We propose an accurate and efficient post-training quantization framework for pre-trained diffusion models that preserve the generation performance in image generation with 6-bit weights and activations.
    \item We present the distribution-aware quantization and activate timestep selection function to significantly reduce quantization errors across generation timesteps and search the representative calibration images according to structural risk minimization principle, so that the rounding functions can be optimized with more informative supervision.
    \item We conduct extensive experiments on a wide variety of datasets for image generation, and the results clearly demonstrate the superiority of the presented method.
\end{itemize}

\begin{figure*}[t]
	\centering
        \includegraphics[width=0.98\linewidth]{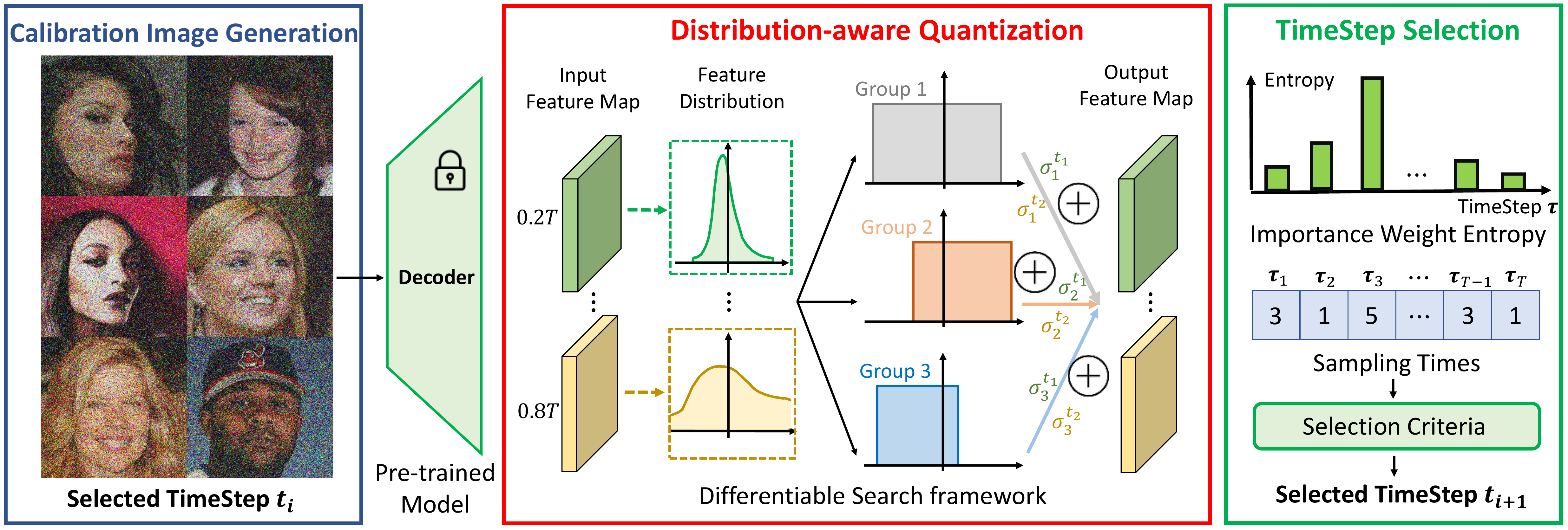}
        \vspace{-0.26cm} 
	\caption{The overall pipeline of our method. The calibration images are generated according to the selected timesteps, and activations in the pre-trained diffusion models are parallelly quantized by rounding functions of all groups. The output feature maps are acquired by adding the quantized value with the importance weights, where the quantization parameters and the importance weights are jointly optimized. The importance weight entropy and the sampling times are considered in the timestep selection criteria to decide the optimal timestep for calibration image generation in the next round.}
	\vspace{-0.32cm}    
	\label{pipeline}
\end{figure*}

\section{Related Work}
\textbf{Efficient diffusion models: }Diffusion models achieve more satisfying quality and diversity in image generation compared with GANs, while the generation efficiency is significantly decreased due to the iterative noise evaluation process with long timesteps. The denoising diffusion probabilistic model (DDPM) \cite{ho2020denoising} leverages a forward pass for noise perturbation and a reverse process for image denoising. Existing methods mainly focus on leveraging a shorter sampling path without sizable performance degradation, which can be divided into two categories including convergence speedup and sampling path selection. Convergence speedup methods aim to discretize the stochastic differential equations (SDE) or the ordinary differential equations (ODE) with minimized discretization errors. Song \emph{et al.} \cite{song2020denoising} modeled the diffusion model with a non-Markov process that considers the original images for noise perturbation, where the convergence for image generation speeds up sizably. Bao \emph{et al.} \cite{bao2022analytic} formulated an analytic form of variance and KL divergence based on a pre-trained score-based model that simultaneously enhanced the log-likelihood and the generation speed. Sampling path selection usually chooses partial timesteps in the denoising process regarding the learning objectives. Watson \emph{et al.} \cite{watson2021learning} searched the best $K$ sampling timesteps for noise evaluation via dynamic programming, where the goal was to maximize the evidence lower bound (ELBO) in the reverse process. Due to the inconsistent performance between the training ELBO and the generation quality, they further presented Kernel Inception Distance (KID) \cite{watson2022learning} as the optimization objective to differentiably search the sampling timesteps. In this paper, we aim to reduce the complexity of single-step denoising process by quantization, which is orthogonal to the acceleration techniques of sampling path shortening.

\textbf{Network quantization: }Network quantization has aroused extensive interest in computer vision due to the significant reduction in storage and computational cost, as the full-precision variables are substituted by quantized values and the multiply-add (MAC) operations are replaced by integer arithmetics. Quantization-aware training (QAT) \cite{liu2020bi,choi2018pact} finetunes the quantized network with training dataset of full-precision models. Due to the inaccessibility of the full training set and the extremely high training cost, post-training quantization (PTQ) \cite{liu2021post,nagel2020up,fang2020post,li2021brecq} that optimizes the rounding functions with a small calibration set is more practical in realistic applications. Choukroun \emph{et al.} \cite{choukroun2019low} minimized the $l_2$ distance between the quantized and full-precision tensors to avoid obvious task performance degradation, and Zhao \emph{et al.} \cite{zhao2019improving} duplicated the channels with outliers and halved the value so that the clipping loss could be reduced without increasing the rounding errors. Liu \emph{et al.} \cite{liu2021post} preserved the relative ranking orders of the self-attention in vision transformers to prevent information loss in post-training quantization, and explored mixed-precision quantization strategy according to the nuclear norm of attention map and features. Zero-shot PTQ further extends the limits that efficiently quantize neural networks without any real image data. Cai \emph{et al.} \cite{cai2020zeroq} optimized the pixel values of the generated images to enforce the statistics of sample batches to mimic the batch normalization (BN) layers in the full-precision networks. Li \emph{et al.} \cite{li2022patch} further extended PTQ framework to transformer architectures by diversifying the self-attention of different patches with patch similarity metrics. As Shang \emph{et al.} \cite{shang2022post} and Li \emph{et al.} \cite{li2023q} observed, different activation distribution across timesteps and the effectiveness change of calibration images acquired in various timesteps amplify the quantization errors in existing methods. To avoid the overfitting of the step-wise quantization caused by limited calibration samples, we present the distribution-aware quantization for diffusion models across timesteps with significantly reduced learnable parameters. Meanwhile, different from \cite{shang2022post} that manually assigned the timestep index for calibration generation, we generalize the structural risk minimization principle to discover the optimal timesteps.

\section{Approach}
In this section, we first introduce the preliminaries of post-training quantization for diffusion models and then detail the distribution-aware quantization across generation timesteps with the differentiable search framework. Finally, we demonstrate the timestep selection for calibration image generation according to the structural risk minimization principle.

\subsection{Network Quantization}
Diffusion models leverage a forward pass to impose noise on images and a reverse pass to transform Gaussian noise into an image for generation. Denoting the real data as $\bm{x}_0$ and the latent image at the $t_{th}$ step as $\bm{x}_t$, the probability of the forward process can be represented as follows:
\begin{equation}
	q(\bm{x}_{t}|\bm{x}_{t-1})=\mathcal{N}(\bm{x}_t|\sqrt{1-\beta_t}\bm{x}_{t-1}, \beta_t\bm{I}),
\end{equation}where $\beta_t$ means the variance schedule at the $t_{th}$ step that indicates imposed Gaussian noise to the latent image. When the total number of forward steps denoted as $T$ becomes large enough, the latent image $\bm{x}_T$ can be regarded as the standard Gaussian noise. We leverage an approximated condition distribution $p_{\theta}(\bm{x}_{t-1}|\bm{x}_{t})$ to generate the latent image in reverse process due to the intractability of true distribution $q(\bm{x}_{t-1}|\bm{x}_{t})$, where the approximated distribution is parameterized by neural networks with weight $\theta$:
\begin{equation}
	p_{\theta}(\bm{x}_{t-1}|\bm{x}_{t})=\mathcal{N}(\bm{x}_{t-1}|\bm{\mu}_{\theta,t}(\bm{x}_t),\bm{\Sigma}_{\theta,t}(\bm{x}_t)).
\end{equation}The training process of diffusion model aims to minimize the negative log-likelihood with the evidence lower bound optimization in variational inference:
\begin{equation}
	L_{VLB}=\mathbb{E}_{q(\bm{x}_{0:T})}[\log\frac{q(\bm{x}_{1:T}|\bm{x}_0)}{p_{\theta}(\bm{x}_{0:T})}]\geqslant-\mathbb{E}_{q(\bm{x}_0)}\log \bm{p}_{\theta}(\bm{x}_0).
\end{equation}In the practical applications, iterative noise estimation process is implemented with the diffusion model for content generation, and the heavy computational cost of the reverse phase disables the deployment in resource-constrained devices such as mobile phones and robots.  To accelerate the denoising process for each reverse step, post-training quantization leverages a small calibration set to learn the rounding function parameters for weights and activations of the decoder, where the quantization function can be represented as follows:
\begin{equation}
\hat{x}=s\cdot\Phi(\left[x/s\right], z_{min}, z_{max}),
\end{equation}where $x$ and $\hat{x}$ represent real-valued and quantized matrix respectively. $[\cdot]$ means the rounding function to the nearest integer and $\Phi$  is the clipping operation that regularizes the element into the range from $z_{min}$ to $z_{max}$. The quantization scaling parameter $s$ indicates the interval between adjacent rounding points. As empirically demonstrated in \cite{shang2022post}, the activation distribution varies significantly across different timesteps during the reverse process, and the shared rounding functions usually cause severe quantization errors for image generation. Moreover, randomly selecting the timestep to generate latent images for calibration set construction fails to provide sufficient information for generalizable quantization function learning.

\subsection{Distribution-aware Quantization for Diverse Activation Distribution}
Since the activation distribution changes significantly across timesteps, discretizing the full-precision intermediate features in similar data distribution with the same quantization functions can reduce the quantization errors. We first describe the distribution-aware quantization scheme and then illustrate the differentiable group assignment of timesteps. 

Shared quantization functions may cause large clipping errors for widely distributed activations and rounding errors for narrowly distributed ones. Directly assigning specific rounding functions for network activations in each timestep leads to overfitting in optimization because of the limited calibration samples, and quantizing activations in the timestep where optimal quantization range is similar with the same rounding functions can achieve better trade-off between the quantization accuracy and rounding function generalizability. Assuming partitioning all $T$ timesteps into $C$ groups, the quantization strategy can be written in the following:
\begin{equation}
\setlength\abovedisplayskip{6pt}
\setlength\belowdisplayskip{9pt}
	\hat{x}=s_{g(i)}\cdot\Phi(\left[x/s_{g(i)}\right], z_{min}^{g(i)}, z_{max}^{g(i)}),
\end{equation}where $g(i)$ represents the assigned group for activations in the $i_{th}$ timestep. Meanwhile, $s_{g(i)}$,  $z_{min}^{g(i)}$ and $z_{max}^{g(i)}$ respectively stand for scale parameters, the lower bound and the upper bound of quantization for activations in the $i_{th}$ timestep. Assigning the optimal group indexes for different timesteps is critical in distribution-aware quantization to reduce the quantization errors without obvious computation overhead. Since enumerating assignment permutation is NP-hard to find the optimal solution, we extend the differentiable search framework to efficiently partition timesteps with minimal quantization errors. In the differentiable search, the latent images are quantized by all quantization functions, whose output values are summed with learnable importance weights to form the input for the next layer in the diffusion model:
\begin{equation}
\setlength\abovedisplayskip{3pt}
\setlength\belowdisplayskip{4.9pt}
	\hat{x}=\sum_{g=1}^{G}\sigma_gs_{g}\cdot\Phi(\left[x/s_{g}\right], z_{min}^{g}, z_{max}^{g}),
\end{equation}where $\sigma_g$ means the importance weight of the quantization function for the $c_{th}$ group with the normalization $\sum_{g=1}^{G}\sigma_g=1$. When the training process completes, the rounding function with the largest importance weight is selected to be the search results for group-wise quantization. Despite the noise estimation loss of diffusion models, we also enforce the importance weights to approach zero or one by minimizing the entropy to avoid discretization errors in rounding function selection. The overall optimization objective $J$ is written as follows, where we denote $\bm{\epsilon}_{\theta}(\sqrt{\overline{\alpha}_t}\bm{x}_0+\sqrt{1-\overline{\alpha}_t}\bm{\epsilon},t)$ as $\bm{\epsilon}_{\theta}$ for simplicity:
\begin{equation}\label{obj}
\setlength\abovedisplayskip{7pt}
\setlength\belowdisplayskip{7pt}
	\min \limits_{t, \bm{x}_0,\bm{\epsilon}}J=J_d+\lambda J_e=||\bm{\epsilon}-\bm{\epsilon}_{\theta}||_2^2+\lambda\sum_{g=1}^{G}-\sigma_g^t\log\sigma_g^t,
\end{equation}

where $J_d$ and $J_e$ respectively represent the simplified variational lower bound of diffusion model objective and the discretization minimization loss in differentiable search, and the hyperparameter $\lambda$ balances the importance of different terms. $\sigma_g^t$ demonstrates the importance weights of the quantization function for the $g_{th}$ group in the $t_{th}$ timestep. The noise $\bm{\epsilon}$ from standard Gaussian distribution is approximated by the predicted noise $\bm{\epsilon}_{\theta}$ in the optimization objective. The diffusion parameter $\overline{\alpha}_t=\prod_{i-1}^{t}1-\beta_i$ controls the strength of noise in diffusion. We jointly update the parameters in quantization functions and the importance weights until convergence or achieving the maximal iteration steps, and the discretized hypernetwork is directly employed to generate images efficiently. 

\subsection{TimeStep Selection for Calibration Generation}
The pipeline of post-training quantization for iterative reverse process in diffusion models differs significantly from that in conventional vision models. Leveraging latent images in all timesteps leads to unbearable training cost for quantization function learning, and latent images in adjacent timesteps can only offer redundant information for parameter optimization. On the contrary, randomly select part of the timesteps usually fails to provide sufficient supervision that is representative to demonstrate the real distribution of the latent images. Therefore, it is desirable to actively sample the timesteps to generate latent images for quantization parameter learning with effective guidance. We generate representative samples by structural risk minimization, which minimize the distance between selected and real distribution. We first introduce the extension of SRM principle to active timestep selection, and then formulate the selection criteria that can be feasibly computed.

Structural risk minimization principle minimizes the upper bound of the true risk on unseen data distribution, where the bound can be written as follows for a dataset containing $n$ samples with the probability at least $1-\delta$\cite{bartlett2002rademacher}:
\begin{equation}
\setlength\abovedisplayskip{6pt}
\setlength\belowdisplayskip{8pt}
    E(J(x))\leqslant\overline{E(J(x))}+2R_n(\mathcal{F})+\sqrt{\frac{\ln 1/\delta}{n}},
\end{equation}where $E(J(x))$ and $\overline{E(J(x))}$ respectively illustrate the true expectation of the risk $J$ for real data distribution $x$ and the empirical expectation of that for sampled data from $x$, and $R_n(\mathcal{F})$ is the Rademacher complexity over the function class $\mathcal{F}$. The SRM principle requires the data to be sampled from i.i.d. distribution, while the latent images in selected timesteps should be more informative and representative. Therefore, we rewrite the SRM principle in the following way, where the detailed formulation is in the supplementary:
\begin{equation}
	E(J)\leqslant\overline{E_S(J)}+MMD(p(X),p(X_s))+\mathcal{R}_0,
\end{equation}where we omit the data distribution $x$ for simplicity. $\overline{E_S(J)}$ denotes the empirical risk of the latent images of selected timesteps for noise estimation, and $\mathcal{R}_0$ demonstrates the complexity of the diffusion model in the reverse process. $X$ and $X_s$ stand for the distribution of latent images generated in all timesteps and the selected ones. The maximal mean discrepancy between two distributions $p(X)$ and $p(X_s)$ is represented as $MMD(p(X),p(X_s))$, which demonstrates the generalization ability of the calibration sets for quantization learning. The first criteria acquired by worst-case empirical risk of sampled latent images is formulated as follows for timesteps selection :
\begin{equation}
	\min\limits_{t, \bm{x}_t}\overline{E_S(J)}=\sum_{\bm{x}_t\in\mathcal{S}}J_d+\lambda J_e,
 \setlength\belowdisplayskip{2pt}
\end{equation}
where $\mathcal{S}$ represents the images selected in the calibration set, and $J$ is the optimization objective defined in (\ref{obj}). This formula aims to train the highly quantified diffusion model with the constructed calibration sets. Since the original latent $\bm{x}_T$ can be regarded as the standard Gaussian noise without bias, the objective $J_d$ does not affect the worst-case empirical risk with different timesteps. The variance of $J_e$ influences the worst-case empirical risk across timesteps, because the entropy of the importance weights of distribution-aware quantization functions changes with the timesteps. Therefore, the criteria $s_1$ from the empirical risk minimization can be transformed to selecting the timestep with the highest entropy of importance weights as $s_1=\sum_{g=1}^{G}-\sigma_g^t\log\sigma_g^t$. 

Meanwhile, The definition of maximal mean discrepancy can be written as follows, where we denote $MMD(p(X),p(X_s))$ as $M$ for simplicity:
\begin{equation}\label{MMD}
\setlength\abovedisplayskip{6pt}
\setlength\belowdisplayskip{7pt}
    \begin{split}
	\min_{t} M&=\sup|| \frac{1}{|U|}\sum_{\bm{x}_t\in U}\bm{\epsilon}_{\theta}-\frac{1}{|\mathcal{S}|}\sum_{\bm{x}_t\in\mathcal{S}}\bm{\epsilon}_{\theta}||\\
        &= \frac{\varphi}{N_t+1} \propto \frac{1}{N_t+1},
    \end{split}
\end{equation}
where $U$ means the full set containing all original latent and timesteps for calibrating image selection, and $|\cdot|$ represents the number of elements in the set. $\varphi$ is a constant in timestep sampling and $N_t$ denotes the number of sampling times for the $t_{th}$ timestep in calibration set construction. Because the estimated noise of the samples in the full set is intractable, we utilize the number of sampling times to optimize the maximal mean discrepancy based on upper confidence bound (UCB) principle \cite{auer2002finite} which achieves exploitation-exploration trade-off when sampling. A detailed derivation of the formula (\ref{MMD}) can be found in the supplementary material. For the timestep when we sample a large number of latent images for calibration set construction, the maximal mean discrepancy becomes low as we acquire sufficient information of the latent image distribution in this timestep. Therefore, we explore latent images in the timestep with few sampling times to further minimize the maximal mean discrepancy with high marginal benefits. The criteria $s_2$ from the maximal mean discrepancy is designed as $s_2=1/(N_t+1)$. The overall timestep selection criteria can be written as follows:
\begin{equation}
\setlength\abovedisplayskip{3pt}
\setlength\belowdisplayskip{3pt}
	\max \limits_{t} s=s_1+\eta s_2=\sum_{g=1}^{G}-\sigma_g^t\log\sigma_g^t+\frac{\eta}{N_t+1},
\end{equation}where $\eta$ is a hyperparameter to balance the importance of empirical risk and maximal mean discrepancy. The overall pipeline of our method is depicted in Figure \ref{pipeline}. With the optimally selected timesteps, the generated calibration images can provide effective supervision for quantization function learning, which can be well generalized in deployment.

\section{Experiments}
In this section, we first introduce the implementation details of our method. We then conduct ablation studies to evaluate the effectiveness of the distribution-aware quantization and the optimal timestep selection for calibration image generation. Meanwhile, we visualize the importance evolution during the differentiable search and analyze the influence of hyperparameters on generation quality. Finally, we compare our method with the state-of-the-art post-training quantization frameworks in diffusion models to show our superiority.

\subsection{Implementation Details}
We utilize the diffusion frameworks for post-training quantization including DDIM \cite{song2020denoising} and LDMs \cite{rombach2022high} with their pre-trained weights, which require 100 iterative denoising timesteps for image generation in most experiments. We set the bitwidth of quantized weight and activation to 6 and 8 to evaluate our method in different quality-efficiency trade-offs and utilized the uniform quantization scheme where the interval between adjacent rounding points was equal. For distribution-aware quantization across different timesteps, we partitioned all timesteps into eight groups in most experiments. We followed the initialization of the quantization function parameters in \cite{li2023q} for the baseline methods and our APQ-DM, where we minimized the $l_p$ distance \cite{nahshan2021loss, wei2022qdrop} between the full-precision and quantized activations to optimize the value range for clipping. The hyperparameters $\lambda$ in the objective of differentiable search and $\eta$ in the timestep selection criteria were set to 0.8 and 1.5 respectively.

For the parameter learning in differentiable search, we generated 1024 images for hyper-network learning where the batchsize was assigned with 64 for calibration set construction. The learning rate was initialized to $3e^{-3}$ and $5e^{-3}$ for 6 and 8-bit diffusion models and ended up with $1e^{-5}$ for all bitwidth settings with 0.05 decaying strategy. The quantization function parameters and the importance weights were jointly updated for 10 epochs in the differentiable search, and the acquired distribution-aware quantization function was directly employed for image generation.

\begin{table}
    \centering
    \footnotesize
    \renewcommand\arraystretch{1.2}
    \vspace{-0.1cm}
	\begin{tabular}{p{1.1cm}<{\centering}|p{0.9cm}<{\centering}|p{1.0cm}<{\centering}p{1.0cm}<{\centering}p{0.9cm}<{\centering}p{0.9cm}<{\centering}}
         \hline
	Bitwidth & Group & C-Error & G-Error & IS$\uparrow$ & FID$\downarrow$\\
	\hline
	\multirow{4}{*}{W8A8} & 1 & 1.16 & 1.22 & 8.93 & 5.32 \\
	&4 & 0.87 & 0.93 & 8.97 & 4.73 \\
	&8 & 0.79 & 0.82 & 9.07 & 4.24 \\
	&16 & 0.75 & 0.86 & 8.98 & 4.29 \\
	\hline
	\multirow{4}{*}{W6A6} & 1 & 1.92 & 2.03 & 8.82 & 9.73 \\
	&4 & 1.88 & 1.76 & 8.92 & 7.10 \\
	&8 & 1.57 & 1.68 & 9.06 & 6.57 \\
	&16 & 1.52 & 1.74 & 9.24 & 6.77 \\
		\hline
	\end{tabular}
        \vspace{-0.2cm}
	\caption{Effects of the number of timestep partitions in the distribution-aware quantization. C-Error and G-Error depict the quantization errors of activations in calibration and generation respectively.}
	\label{ablation group number}
\vspace{-0.3cm}
\end{table}

\begin{table}
    \centering
    \footnotesize
    \renewcommand\arraystretch{1.34}
    \vspace{-0.1cm}
	\begin{tabular}{p{1.1cm}<{\centering}|p{1.1cm}<{\centering}|p{0.9cm}<{\centering}p{0.9cm}<{\centering}p{0.9cm}<{\centering}p{0.9cm}<{\centering}}
		\hline
	\multirow{2}{*}{Method} & \multirow{2}{*}{Bitwidth} & \multicolumn{4}{c}{Size of Calibration Set}\\
        \cline{3-6}
        && 128 & 256 & 512 & 1024 \\
	\hline
	\multirow{2}{*}{Random} 
	& W8A8 & 5.72 & 5.64 & 5.34 & 5.41\\
	& W6A6 & 11.65 & 10.12 & 9.18 & 8.92\\
	\hline
	\multirow{2}{*}{Heuristic} 
	& W8A8 & 5.75 & 5.56 & 5.27 & 5.21\\
	& W6A6 & 12.26 & 10.19 & 9.03 & 8.83\\
 	\hline
	\multirow{2}{*}{Active} 
	& W8A8 & 5.99 & 4.46 & 4.49 & 4.24\\
	& W6A6 & 12.61 & 11.73 & 7.83 & 6.57\\
        \hline
	\end{tabular}
        \vspace{-0.2cm}
	\caption{Different timestep sampling strategies for calibration set construction across various sizes of calibration images. WBAB depicts the weights and activations are quantized to B-bit.}
	\label{ablation calibrate set}
 \vspace{-0.3cm}
\end{table}

\begin{figure}[t]
  \centering
  \begin{subfigure}{0.492\linewidth}
    \includegraphics[width=1\linewidth]{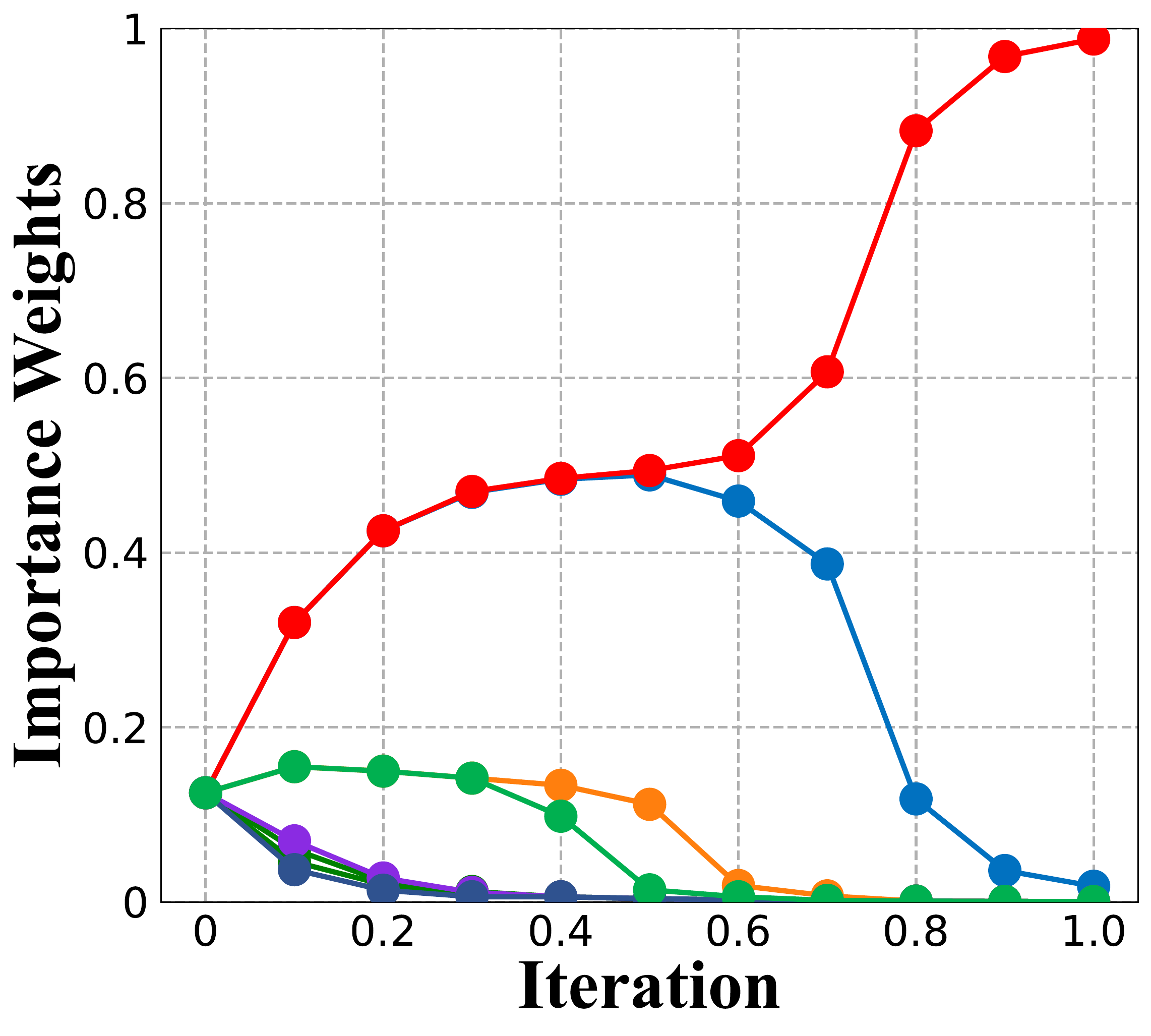}
    \caption{}
    \vspace{-0.2cm}
    \label{importance weight}
  \end{subfigure}
  \begin{subfigure}{0.492\linewidth}
    \includegraphics[width=1\linewidth]{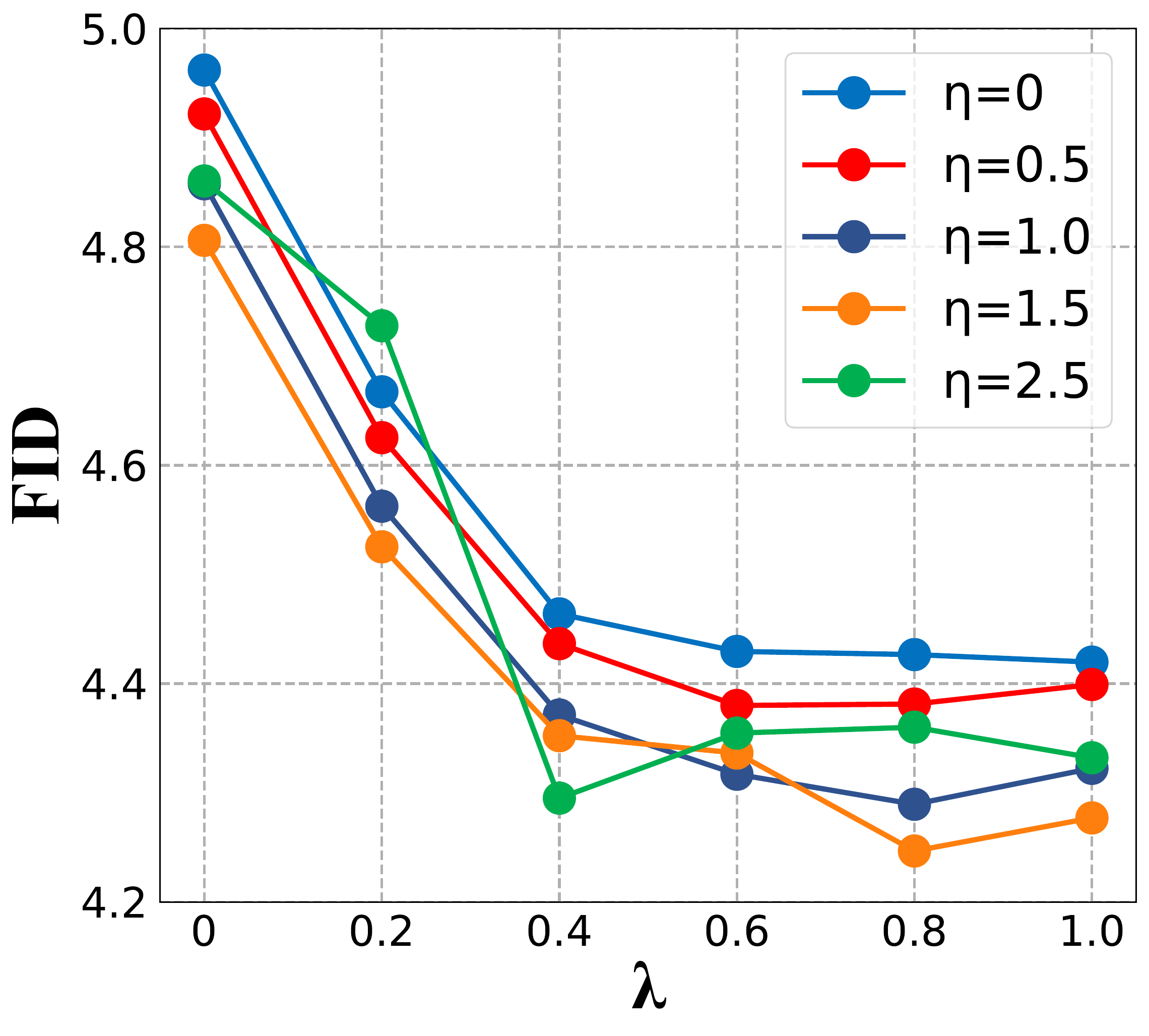}
    \caption{}
    \vspace{-0.2cm}
    \label{ablation hyperparameters}
  \end{subfigure}
  \vspace{-0.4cm}
  \caption{(a) The evolution of branch importance weights during the differentiable search. (b) The generation quality w.r.t. different hyperparameters $\lambda$ and $\eta$.}
  \label{fig:3}
  \vspace{-0.3cm}
\end{figure}

\begin{table*}
    \centering
    \vspace{0.1cm}
    \renewcommand\arraystretch{1.3}
    \footnotesize
    \begin{tabular}{p{1.7cm}<{\centering}|p{1.2cm}<{\centering}|p{0.7cm}<{\centering}p{0.7cm}<{\centering}p{0.7cm}<{\centering}|p{0.7cm}<{\centering}p{0.7cm}<{\centering}p{0.7cm}<{\centering}|p{0.7cm}<{\centering}p{0.7cm}<{\centering}p{0.7cm}<{\centering}|p{0.7cm}<{\centering}p{0.7cm}<{\centering}p{0.7cm}<{\centering}}
        \hline
        \multirow{2}{*}{Method} & \multirow{2}{*}{Bitwidth} &  \multicolumn{3}{c|}{Cifar-10} & \multicolumn{3}{c|}{CelebA} & \multicolumn{3}{c|}{LSUN-Bedroom} & \multicolumn{3}{c}{LSUN-Church} 
        \\
        \cline{3-14}
        &&IS$\uparrow$&FID$\downarrow$&sFID$\downarrow$&IS$\uparrow$&FID$\downarrow$&sFID$\downarrow$&IS$\uparrow$&FID$\downarrow$&sFID$\downarrow$&IS$\uparrow$&FID$\downarrow$&sFID$\downarrow$ \\
        \hline
        
        Baseline & FP & 9.07 & 4.23 & 4.37 & 2.61 & 6.49 & 13.82 & 2.45 & 6.39 & 9.45 & 2.76 & 10.98 & 16.16 \\
            \hline
            LSQ & \multirow{4}{*}{W8A8} &8.74&13.78& 6.93 &2.29&15.02& 15.99 &2.13&16.95& 18.85 &2.58&28.49& 30.95 \\
           PTQ4DM &  & 8.82 & 5.69 & 4.42 & 2.43 & 6.44 & 14.15 & 2.23 & 7.48 & 12.42 & 2.76 & 10.98 & 17.28 \\
            Q-Diffusion & & 8.87 & 4.78 & 4.49 & 2.41 & 6.60 &14.19 & 2.27 & 7.04 & 12.24 & 2.72 & 12.72 & 16.96\\
        APQ-DM & & \textbf{9.07} & \textbf{4.24} & \textbf{4.37} & \textbf{2.58} & \textbf{6.07} & \textbf{13.09} & \textbf{2.55} & \textbf{6.46} & \textbf{11.82} & \textbf{2.84} & \textbf{9.04} & \textbf{16.74}\\
           \hline
           LSQ & \multirow{4}{*}{W6A6} &8.34&35.96 & 37.04 &1.94&78.37& 85.04&1.68&122.45& 126.24&1.87&131.78& 140.39\\
           PTQ4DM &  & 8.72 & 11.28 & 6.92 & 2.13 & 24.96 & 20.95 & 2.11 & 16.85 & 19.65 & 2.48 & 32.85& 36.74 \\
            Q-Diffusion & &  8.76 & 9.19 & 5.80 & 2.16 & 23.37 & 19.83 & 2.09 & 17.57 & 18.74 & 2.52 & 33.77 & 35.27\\
        APQ-DM & &  \textbf{9.06} & \textbf{6.57} & \textbf{4.71} & \textbf{2.30} & \textbf{16.86} & \textbf{17.85} & \textbf{2.30} & \textbf{15.72} & \textbf{17.24} & \textbf{2.63} & \textbf{24.75} & \textbf{29.24}\\
        \hline
    \end{tabular}
    \vspace{-0.25cm}
    \caption{Comparisons with the state-of-the-arts data-free post-training quantization methods on unconditional image generation for DDIM diffusion models across various datasets and bitwidth setting.}
    \label{tab:DDIM}
    \vspace{-0.45cm}
\end{table*}

\subsection{Ablation Study}
In order to investigate the influence of the distribution-aware quantization for network activations across different timesteps, we vary the number of groups with different trade-offs between quantization accuracy and rounding function generalizability. To show the effectiveness of our active timestep selection for calibration set generation, we compare our strategy with various sampling techniques. Meanwhile, we modified the hyperparameter $\lambda$ and $\eta$ to demonstrate the effect of the discretization loss in rounding function selection and the maximal mean discrepancy in timestep selection criteria. All experiments in the ablation study were conducted with the $32\times32$ cifar-10 dataset and the DDIM diffusion framework.

\textbf{Performance w.r.t. the number of timestep groups: }Dividing the timesteps into more groups can significantly reduce the clipping and rounding errors for differently distributed activations in quantization function learning, while may resulting in the rounding function overfitting due to the limited calibration samples and large-scale learnable parameters. Table \ref{ablation group number} illustrates the quantization errors, Inception Score (IS) and FID score for our method that partitions the timesteps into different numbers of groups. Observing the FID and IS for different group partition settings, we conclude that dividing the timesteps into several groups outperforms both the shared quantization policy and the step-wise rounding functions. Therefore, we assign the number of groups for the timestep partition to 8 in the rest of experiments to achieve the optimal trade-off between the quantization accuracy and rounding function generalizability.

\textbf{Performance w.r.t. different timestep sampling strategies for calibration set construction: }We compare our active timestep sampling strategy for calibration set generation with random and heuristic sampling policies \cite{shang2022post}. Random sampling assigns an integer number drawn from uniform distribution from zero to the maximal time steps $T$, and heuristic sampling determines the timestep from a Gaussian distribution $\mathcal{N}(\mu, \frac{T}{2})$ where $\mu$ is less than $\frac{T}{2}$. Table \ref{ablation calibrate set} shows the generation quality for different timestep sampling methods across various sizes of the calibration set. Our active sampling strategy outperforms the random and heuristic sampling policies by a large margin, and the advantage is more obvious for calibration sets with small sizes because informative samples are extremely important for post-training quantization in the scenario without sufficient images.

\textbf{Visualization of importance weight in differentiable search: }Figure \ref{importance weight} depicts the evolution of importance weights during the differentiable search for group assignment, where different colors represent disparate groups. At the early stage of the differentiable search, the differences between importance weights are not obvious because of insufficient calibration images. When the network gradually converges, the principal impacts on the performance result from the quantization functions with different rounding and clipping errors and differentiate group assignments between different data distributions.

\textbf{Performance w.r.t. hyperparameters $\lambda$ and $\eta$: }The hyperparameter $\lambda$ controls the importance of the discretization loss in distribution-aware quantization function in the objective of differentiable search, and $\eta$ balances the empirical risk and the maximal mean discrepancy in the timestep selection. Figure \ref{ablation hyperparameters} depicts the FID for different hyperparameter settings, where the medium value for both parameters achieves the highest generation quality. The model performance is more sensitive to the hyperparameter $\lambda$ because the importance weights of quantization functions in different groups usually have similar distribution as one-hot vector, and slight change to $\lambda$ leads to large perturbation to the overall objective in differentiable search due to the logarithm.

\begin{table*}
    \vspace{0.1cm}
    \centering
    \renewcommand\arraystretch{1.3}
    \footnotesize
    \begin{tabular}{p{1.7cm}<{\centering}|p{1.2cm}<{\centering}|p{0.7cm}<{\centering}p{0.7cm}<{\centering}p{0.7cm}<{\centering}|p{0.7cm}<{\centering}p{0.7cm}<{\centering}p{0.7cm}<{\centering}|p{0.7cm}<{\centering}p{0.7cm}<{\centering}p{0.7cm}<{\centering}|p{0.7cm}<{\centering}p{0.7cm}<{\centering}p{0.7cm}<{\centering}}
        \hline
        \multirow{2}{*}{Method} & \multirow{2}{*}{Bitwidth} & \multicolumn{3}{c|}{CelebA-HQ (U)} & \multicolumn{3}{c|}{Bedroom (U)} & \multicolumn{3}{c|}{Church (U)} & \multicolumn{3}{c}{ImageNet (C)}
        \\
        \cline{3-14}
        &&IS$\uparrow$&FID$\downarrow$&sFID$\downarrow$&IS$\uparrow$&FID$\downarrow$&sFID$\downarrow$&IS$\uparrow$&FID$\downarrow$&sFID$\downarrow$&IS$\uparrow$&FID$\downarrow$&sFID$\downarrow$ \\
        \hline
        Baseline & FP & 3.27 & 6.08 & 9.36 & 2.29 & 3.43 & 7.68 & 2.70 & 4.08 & 10.99 & 180.84 & 11.89 & 6.86\\
            \hline
            LSQ & \multirow{4}{*}{W8A8} & 3.01 & 9.75 & 11.04 & 2.13 & 8.11& 11.40&2.50&7.10& 11.21&154.06&13.26& 22.87\\
           PTQ4DM &  & 3.11 & 8.57 & 10.36 & 2.21 & 4.75 & 10.76 & 2.52 & 5.29 & 12.49 & 161.75 & 12.59 & 13.53\\
             Q-Diffusion& & 3.08 & 8.61 & 10.43 & 2.19 & 4.67 & 10.51 & 2.53 & 4.87 & 12.95 & 166.05 & 12.78 & 12.21\\
        APQ-DM & & \textbf{3.22} & \textbf{6.30} & \textbf{9.25} & \textbf{2.35} & \textbf{3.88} & \textbf{8.55} & \textbf{2.69} & \textbf{4.02} & \textbf{10.70} & \textbf{179.13} & \textbf{11.58} & \textbf{6.31}\\
            \hline
            LSQ & \multirow{4}{*}{W6A6} &2.09&129.84& 135.85&1.34&122.45& 148.19&1.82&135.61& 77.77&115.71&40.77& 48.73\\
           PTQ4DM &  & 2.80 & 19.53 & 21.00 & 2.08 & 11.10 & 14.83 & 2.46 & 11.05 & 20.92 & 140.86 & 13.68 & 23.40 \\
            Q-Diffusion & &2.87 & 18.39 & 20.56 & 2.11 & 10.10 & 14.50 & 2.47 & 10.90 & 21.54 & 146.41 & 13.94 & 22.73\\
        APQ-DM & & \textbf{3.09} & \textbf{16.73} & \textbf{18.75} & \textbf{2.27} & \textbf{9.88} & \textbf{13.29} & \textbf{2.67} & \textbf{6.90} & \textbf{13.53} & \textbf{178.64} & \textbf{11.58} & \textbf{7.40}\\
        \hline
    \end{tabular}
    \vspace{-0.23cm}
    \caption{The generation quality on unconditional (U) and class-conditional (C) image synthesis for LDMs diffusion models across different datasets and bitwidths. }
    \label{tab:LDMs}
    \vspace{-0.35cm}
\end{table*}

\begin{figure*}[t]
  \centering
  \begin{subfigure}{0.3\linewidth}
    \includegraphics[width=1\linewidth, height=1\textwidth]{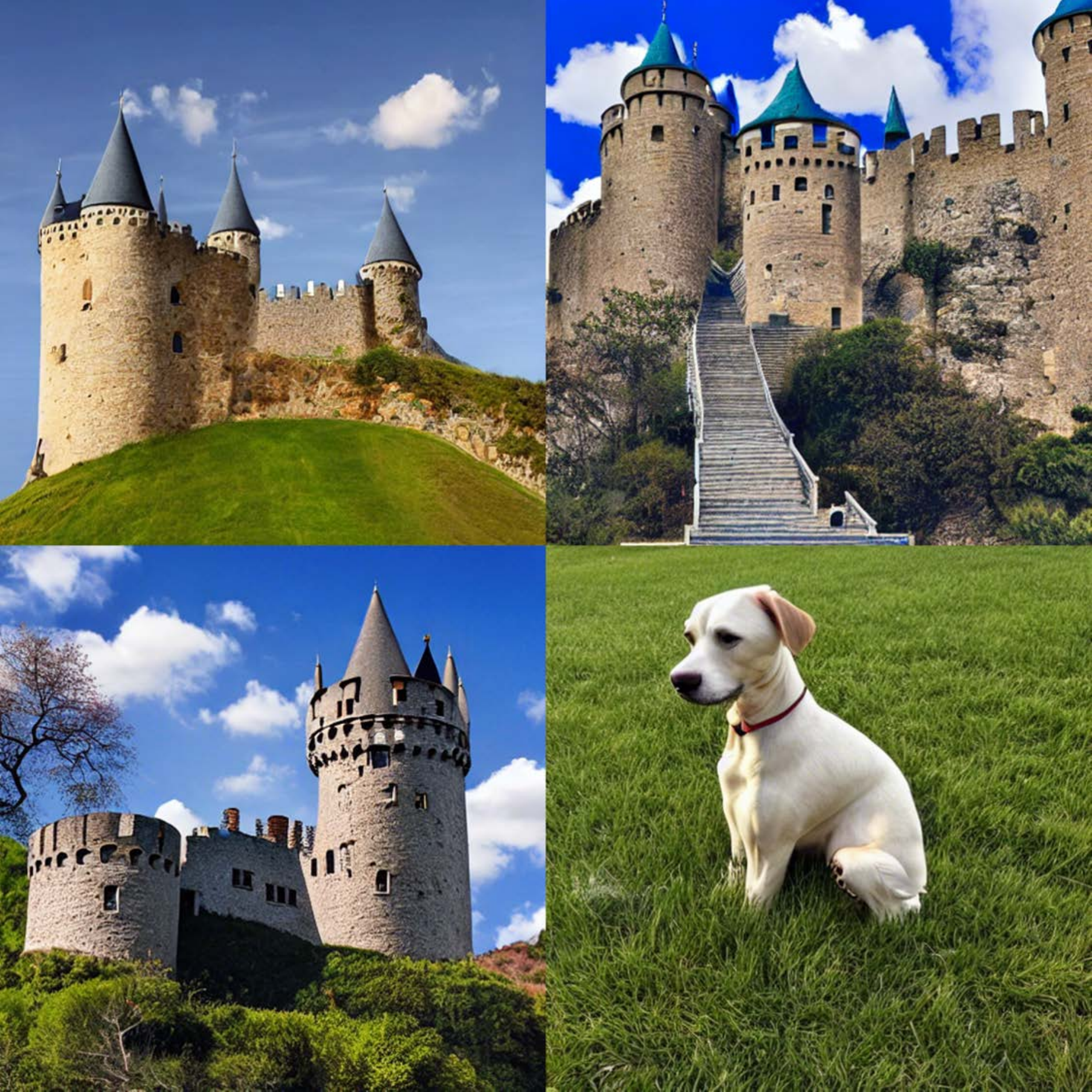}
    \caption{Full Precision}
    \label{fig:4a}
  \end{subfigure}
  \begin{subfigure}{0.3\linewidth}
    \includegraphics[width=1\linewidth, height=1\textwidth]{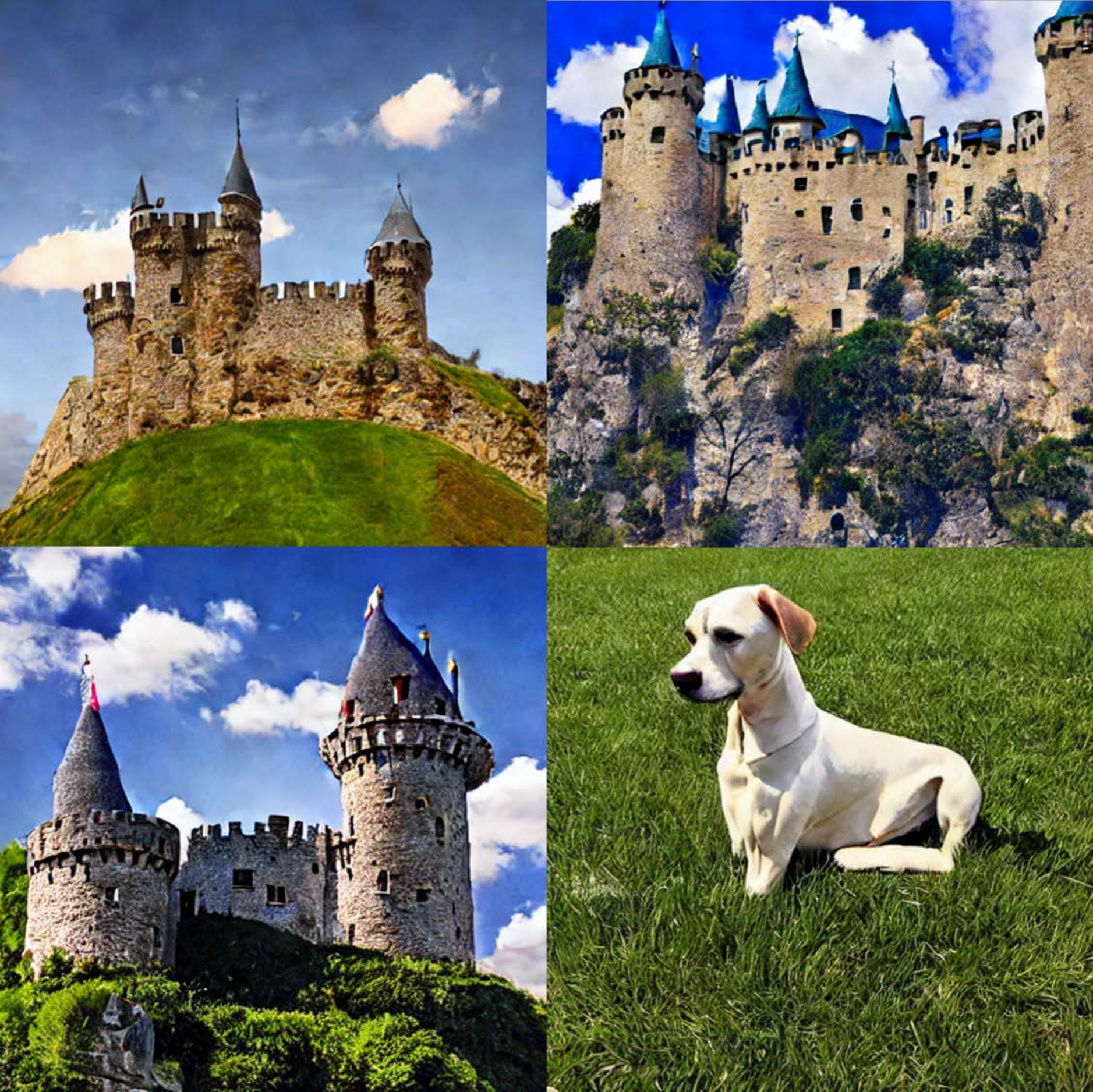}
    \caption{PTQ4DM(6-bit)}
    \label{fig:4b}
  \end{subfigure}
  \begin{subfigure}{0.3\linewidth}
    \includegraphics[width=1\linewidth, height=1\textwidth]{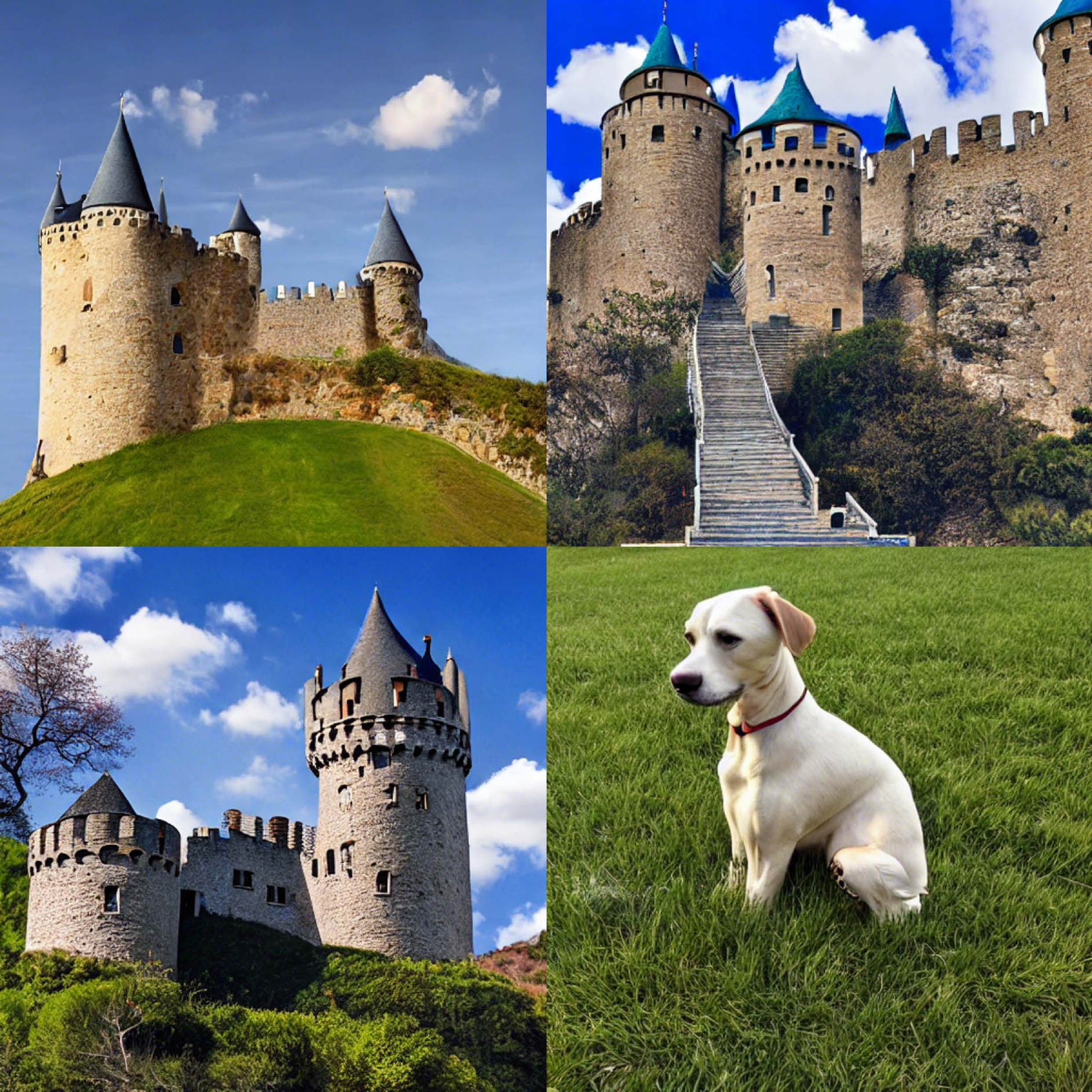}
    \caption{APQ-DM(6-bit)}
    \label{fig:4c}
  \end{subfigure}
  \vspace{-0.33cm}
  \caption{The images generated by quantized Stable Diffusion models and the corresponding text prompts, where different post-training quantization methods are employed.}
  \label{Stable Diffusion}
  \vspace{-0.55cm}
\end{figure*}

\subsection{Comparison with the State-of-the-art Methods}
In this section, we compare our proposed method with the state-of-the-art data-free post-training quantization frameworks including LSQ \cite{esser2019learned} and those specifically designed for diffusion models including PTQ4DM \cite{shang2022post} and Q-diffusion \cite{li2023q}. The IS, FID, and sFID scores of the baseline methods are acquired by implementing the officially released code or our re-implementation. For fair comparison of all listed methods, we leverage the rounding function in LSQ for quantization and de-quantization, and generate latent images with 100 iterative timesteps.

\textbf{Results on unconditional generation: } Unconditional generation samples a random variable for diffusion models to yield images with similar distribution of the training datasets. We evaluate our data-free post-training quantization methods on $32\times32$ Cifar-10 \cite{krizhevsky2009learning}, $64\times64$ CelebA \cite{liu2015deep}, $256\times256$ LSUN-Church Outdoor and LSUN-Bedroom datasets \cite{yu2015lsun} for DDIM frameworks, and evaluate for LDMs diffusion frameworks on $256\times256$ CelebA-HQ \cite{lee2020maskgan}, LSUN-Church Outdoor and LSUN-Bedroom datasets, where the generalization quality and efficiency are reported in Table \ref{tab:DDIM} and Table \ref{tab:LDMs} respectively. LSQ learns the optimal quantization step sizes with minimized discretization errors, while the shared quantization policy across timesteps and randomly constructed calibration set in diffusion models leads to significant quantization loss. PTQ4DM and Q-diffusion employ the step-wise quantization functions to minimize the quantization errors of the diversely distributed activations across timesteps, and presents heuristic timesteps selection criteria for calibration image generation. However, the optimization of large-scale learnable parameters faces the challenges of overfitting due to the very limited quantity of calibration samples, and the data-independent calibration set construction cannot guarantee the optimality of the calibration images. As a result, our method outperforms PTQ4DM by 0.32 (2.55 vs. 2.23) and 1.02 (6.46 vs. 7.48) for IS and FID in LSUN-Bedroom respectively. The computational cost remains the same for baseline methods and our APQ-DM due to the stored rounding parameters. The advantage of our method becomes more obvious for 6-bit diffusion models because quantization errors and calibration sample informativeness are more important for networks with low capacity.

\textbf{Results on conditioned image generation: }
Conditioned image generation synthesizes samples according to text including class names or descriptions. For class-conditional image generation, we discretize the LDMs model that is pre-trained on the $256\times256$ class-conditional ImageNet \cite{deng2009imagenet} dataset, where the guidance strength is set to 3.0 to balance the generation quality and diversity. Table \ref{tab:LDMs} shows the quantitative experimental results for different post-training quantization methods, while our method increases the FID and IS by 1.01 (11.58 vs.12.59) and 17.38 (179.13 vs. 161.75) respectively compared with the state-of-the-art method PTQ4DM. For description-conditional image generation, Figure \ref{Stable Diffusion} demonstrates some examples of the text prompts images that are generated by different quantized Stable Diffusion models, where our method can still acquire plausible images with high-quality details with weights and activations in low bitwidths. Since conditioned image generation is widely adopted in many realistic multimedia applications, our method brings potential to deploy large pre-trained diffusion models on mobile devices or robots under limited resource constraints with satisfying generation quality.

\section{Conclusion}
In this paper, we have presented a novel post-training quantization framework of diffusion model for efficient image generation. We design a differentiable search framework that assigns the optimal partition for each timestep, where network activations are discretized with distribution-aware quantization functions for rounding error minimization. By generalizing the structural risk minimization principle, we select the optimal timesteps for calibration image construction to provide effective supervision in quantization parameter learning. Extensive experiments demonstrate that our methods outperform the state-of-the-art post-training quantization methods across various diffusion architectures. 

\section*{Acknowledgement}
This work was supported in part by the National Natural Science Foundation of China under Grant 62125603, Grant 62321005, and Grant 62336004, and Shenzhen Key Laboratory of Ubiquitous Data Enabling (Grant No. ZDSYS20220527171406015).

{
    \small
    \bibliographystyle{ieeenat_fullname}
    \bibliography{egbib}
}

\clearpage
\setcounter{section}{0}
\renewcommand\thesection{\Alph{section}}

\section{Formulation of (9)}
Structural risk minimization principle minimizes the upper bound of the true risk on unseen data distribution, where the bound can be written as follows for a dataset containing $n$ samples with the probability at least $1-\delta$:
\begin{equation}
	E(J(x))\leqslant\overline{E(J(x))}+2R_n(\mathcal{F})+\sqrt{\frac{\ln 1/\delta}{n}},
\end{equation}where $E(J(x))$ and $\overline{E(J(x))}$ respectively illustrate the true expectation of the risk $J$ for real data distribution $x$ and the empirical expectation of that for sampled data from $x$, and $R_n(\mathcal{F})$ is the Rademacher complexity over the function class $\mathcal{F}$. The SRM principle requires the data to be sampled from i.i.d. distribution, while the latent images in selected timesteps should be more informative and representative. In order to extend the SRM principle in active timestep selection, we omitted $x$ to reformulate the risk bound inequality:
\begin{equation}
	E(J)\leqslant(E(J)-E_T(J))+\overline{E_T(J)}+\mathcal{R}_0,
  \label{con:SRM}
\end{equation}
where $E(J)$ and $E_T(J)$ are the true risk of all latent images and the sampled data. $\mathcal{R}_0 = 2R_n(\mathcal{F})+\sqrt{\frac{\ln 1/\delta}{n}}$ demonstrates the complexity of the diffusion model in the reverse process. In diffusion models, the data $x$ consists of input samples $z$ and target samples $y$, we can rewrite the first term of (\ref{con:SRM}) as follows:
\begin{equation}
    \begin{aligned}
	E(J)-E_T(J)=\int g(z)p(X)dz-\int g(z)p(X_s)dz,
     \label{con:E(J)-E_T(J)}
    \end{aligned}
\end{equation}
where we rewrite $p(z|z\in X)$ and $p(z|z\in X_s)$ as $p(X)$ and $p(X_s)$ respectively for simplicity. $X$ and $X_s$ are the distribution of latent images generated in all timesteps and the selected ones respectively. As $g(z)=\int J\cdot p(y|z)dy$ is bounded and measurable, a bounded and continuous function $\hat{g}(z)$ can guarantee the boundness of (\ref{con:E(J)-E_T(J)}):
\begin{equation}
\setlength\belowdisplayskip{15pt}
    \begin{aligned}
	E(J)-E_T(J)&\leqslant\sup_{\hat{g}(z)}[\int g(z)p(X)dz-\int g(z)p(X_s)dz]\\&=MMD(p(X),p(X_s)),
    \end{aligned}
\end{equation}
where $MMD(p(X),p(X_s))$ represents the maximum mean discrepancy between distribution $p(X)$ and $p(X_s)$. Finally, we rewrite the SRM principle in the following way:
\begin{equation}
\setlength\abovedisplayskip{15pt}
\setlength\belowdisplayskip{15pt}
    E(J)\leqslant\overline{E_T(J)}+MMD(p(X),p(X_s))+\mathcal{R}_0,
\end{equation}where we omit the data distribution $x$ for simplicity. $\overline{E_T(J)}$ denotes the empirical risk of the latent images of selected timesteps for noise estimation. 

\section{Formulation about (11)}

The definition of maximal mean discrepancy can be written as follows, where we denote $MMD(p(X),p(X_s))$ as $M$ for simplicity:
\begin{equation}\label{MMD_s}
\setlength\belowdisplayskip{6pt}
    \begin{split}
	\min_{t} M&=\sup|| \frac{1}{|U|}\sum_{\bm{x}_t\in U}\bm{\epsilon}_{\theta}(\bm{x}_t)-\frac{1}{|\mathcal{S}|}\sum_{\bm{x}_t\in\mathcal{S}}\bm{\epsilon}_{\theta}(\bm{x}_t)||\\
    &=\sup|| E_{\bm{x}_t\in U}(\bm{\epsilon}_{\theta}(\bm{x}_t))-E_{\bm{x}_t\in S}(\bm{\epsilon}_{\theta}(\bm{x}_t))||,
    \end{split}
\end{equation}where $U$ means the full set containing all original latent and timesteps for calibrating image selection, and $|\cdot|$ represents the number of elements in the set. $||\cdot||$ represent for L2 norm calculation. The upper bound of the first term of formula (\ref{MMD_s}) can be written based on the upper confidence bound (UCB) principle as follows:
\begin{equation}\label{UCB}
    \begin{split}
	E_{\bm{x}_t\in U}(\bm{\epsilon}_{\theta}(\bm{x}_t)) = E_{\bm{x}_t\in S}(\bm{\epsilon}_{\theta}(\bm{x}_t)) + \varphi \sqrt{\frac{\ln N}{N_t+1}},
    \end{split}
\end{equation}where $N$ and $N_t$ denote the number of sampling times for the $t_{th}$ timestep and the total sampling times in calibration set construction respectively. $\varphi$ is a constant in timestep sampling to achieve the exploitation-exploration trade-off in UCB principle. $\sqrt{\frac{\ln N}{N_t+1}}$ denotes for uncertainty between the distribution of the full set and the selected samples, which is reduced as the sampling times for the $t_{th}$ timestep raise. 
$N$ in formula (\ref{UCB}) is designed to further explore the timesteps with more uncertainty, when the indeterminacy rises as $t_{th}$ timestep is not selected. However, timestep $t$ is large in diffusion models and the number of selected times $N_t$ is always small for calculating the square root, which leads to large $N$ and instability of uncertainty for the calibration set construction. 
Therefore, we simplify the design of uncertainty and obtain formula (11) in the paper as follows:
\begin{equation}
    \begin{split}
	\min_{t} M = \frac{\varphi}{N_t+1} \propto \frac{1}{N_t+1},
    \end{split}
\end{equation}where we expect to select latent images in the timestep with few sampling times to further minimize the maximal mean discrepancy with high marginal benefits.

\begin{figure*}[t]
  \centering
  \begin{subfigure}{0.33\linewidth}
    \includegraphics[width=1\linewidth, height=1\textwidth]{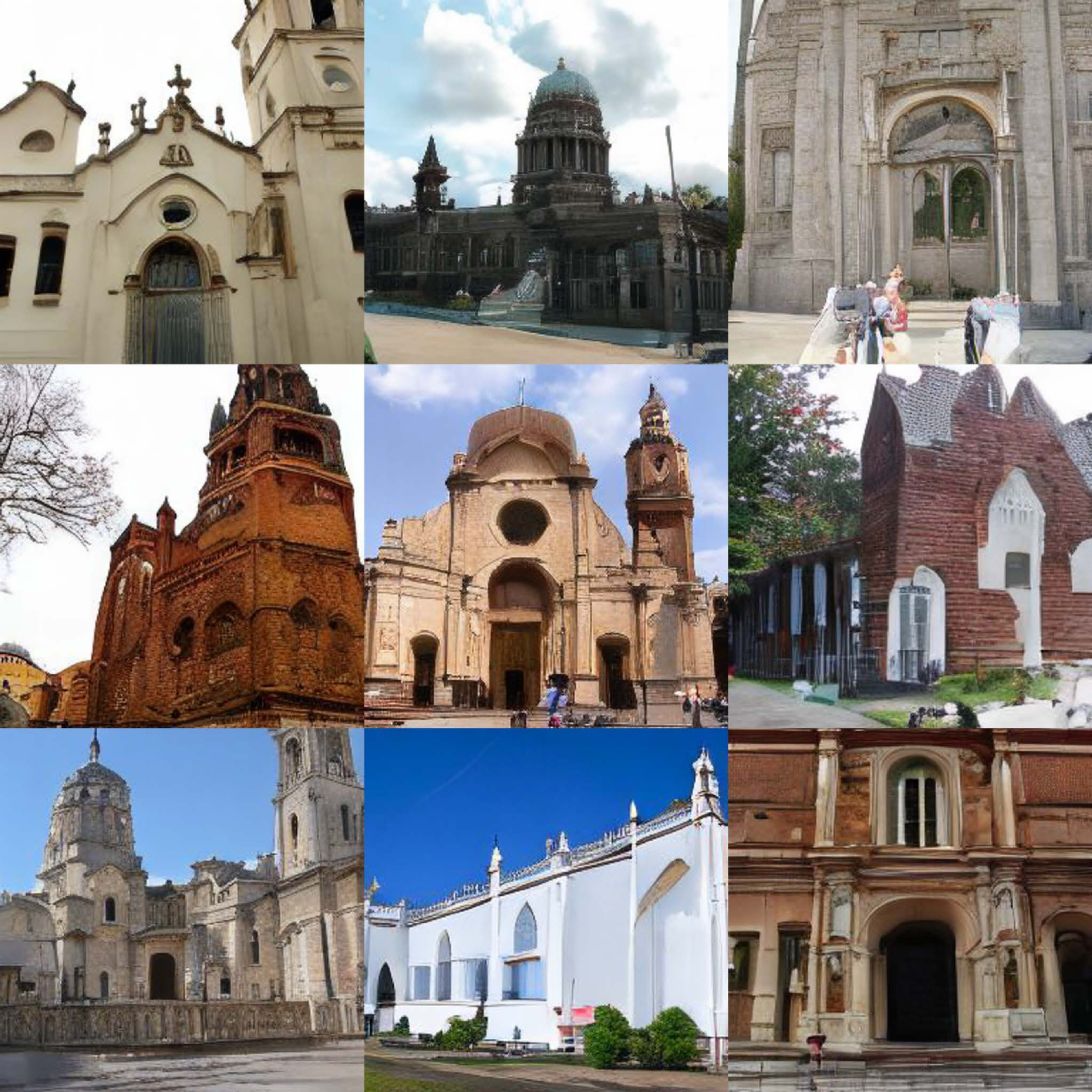}
    \caption{Full Precision}
  \end{subfigure}
  \begin{subfigure}{0.33\linewidth}
    \includegraphics[width=1\linewidth, height=1\textwidth]{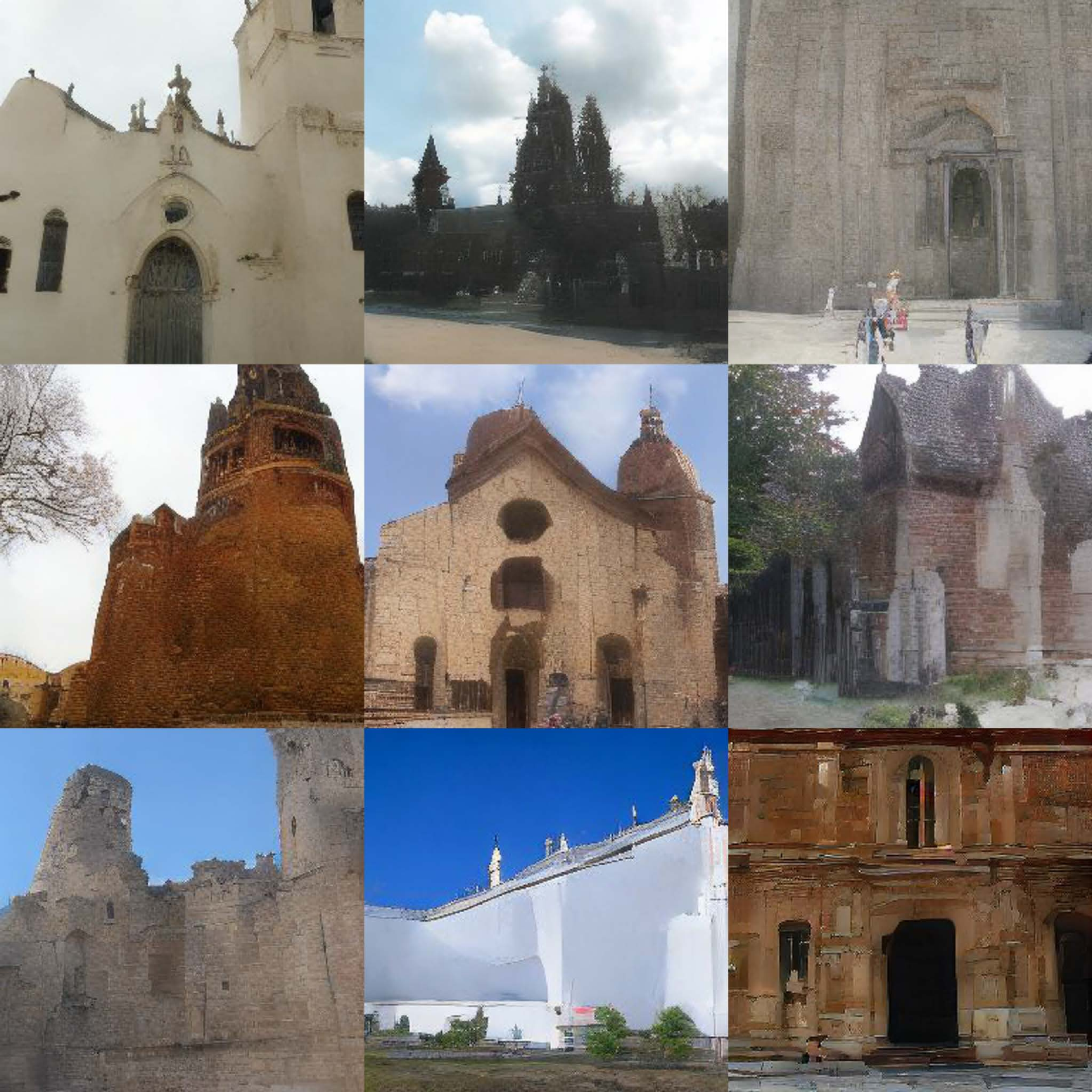}
    \caption{PTQ4DM(6-bit)}
  \end{subfigure}
  \begin{subfigure}{0.33\linewidth}
    \includegraphics[width=1\linewidth, height=1\textwidth]{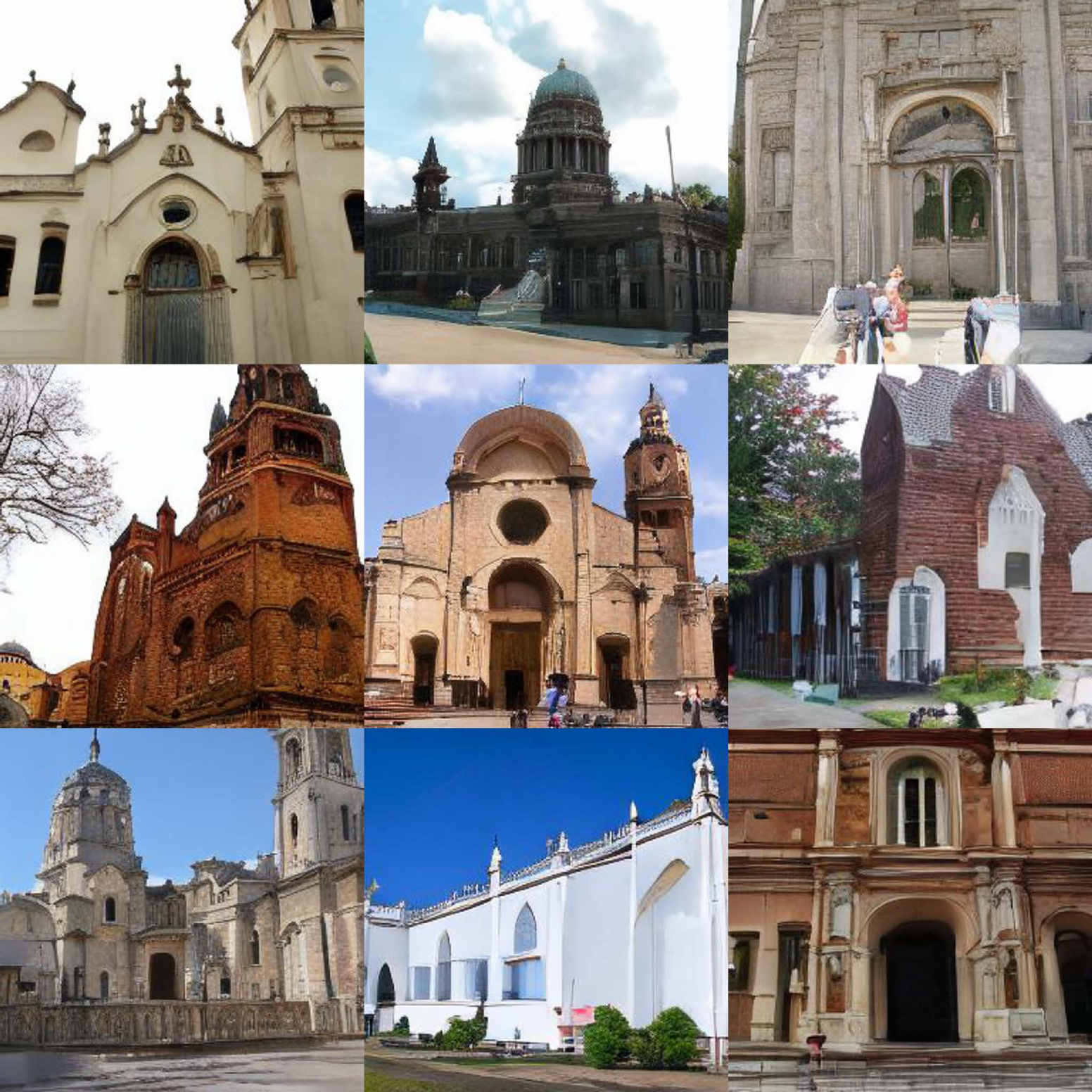}
    \caption{ADP-DM(6-bit)}
  \end{subfigure}
  \caption{$256\times256$ LSUN-Church samples from 100 step LDMs in 6-bit with different post-training quantization methods.}
  \label{Church}
  \vspace{-0.1cm}
\end{figure*}

\begin{figure*}[t]
  \centering
  \begin{subfigure}{0.33\linewidth}
    \includegraphics[width=1\linewidth, height=1\textwidth]{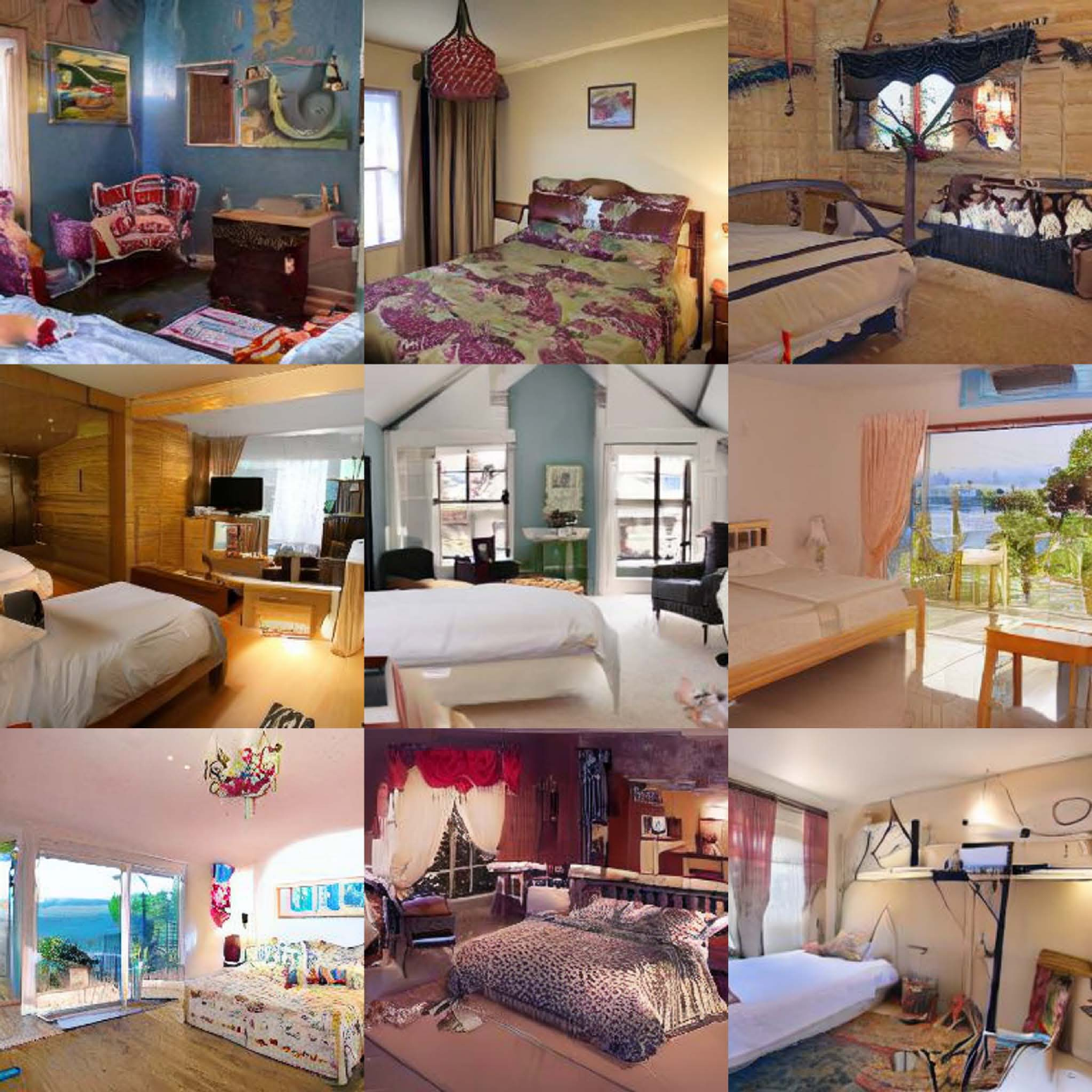}
    \caption{Full Precision}
    \label{fig:4a}
  \end{subfigure}
  \begin{subfigure}{0.33\linewidth}
    \includegraphics[width=1\linewidth, height=1\textwidth]{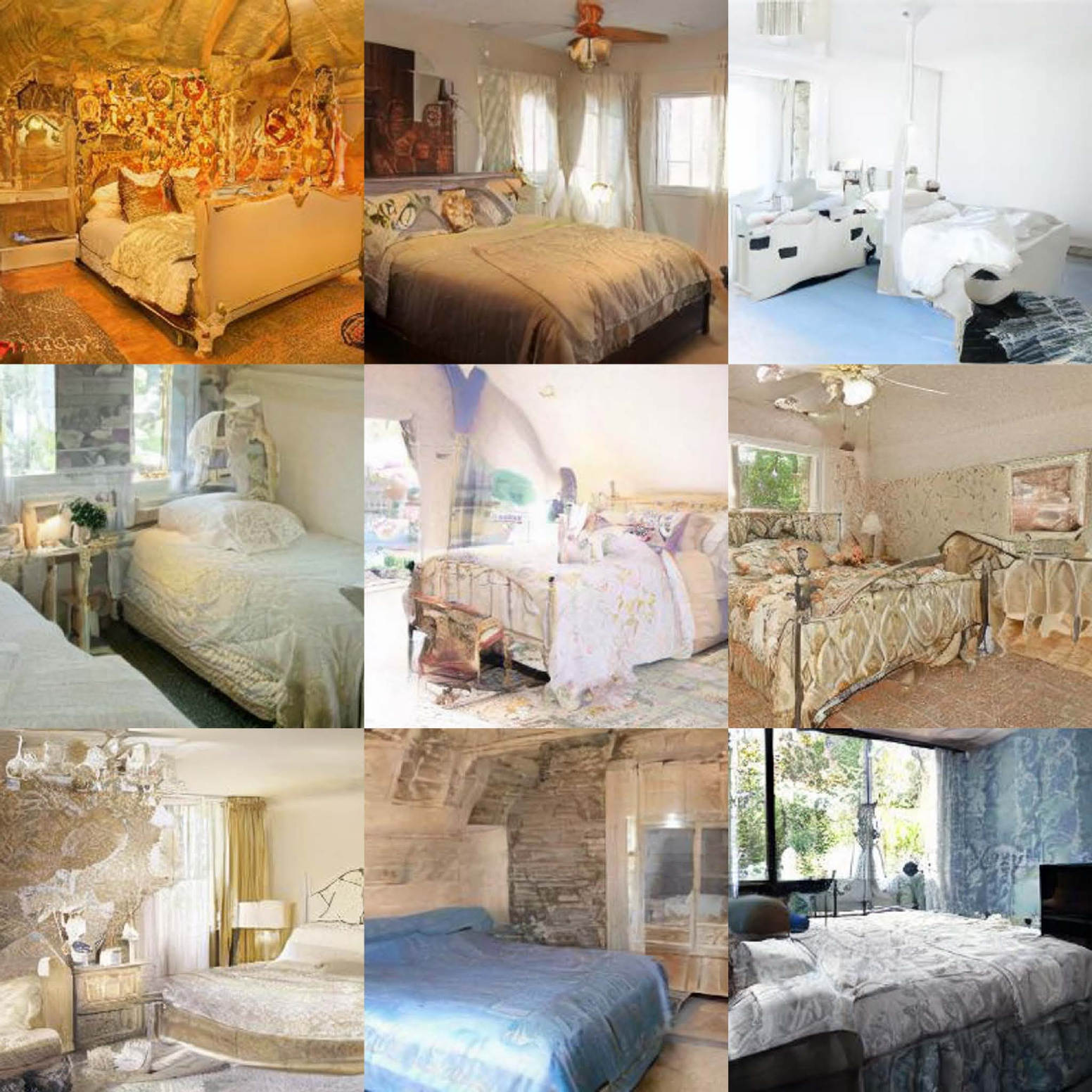}
    \caption{PTQ4DM(6-bit)}
    \label{fig:4b}
  \end{subfigure}
  \begin{subfigure}{0.33\linewidth}
    \includegraphics[width=1\linewidth, height=1\textwidth]{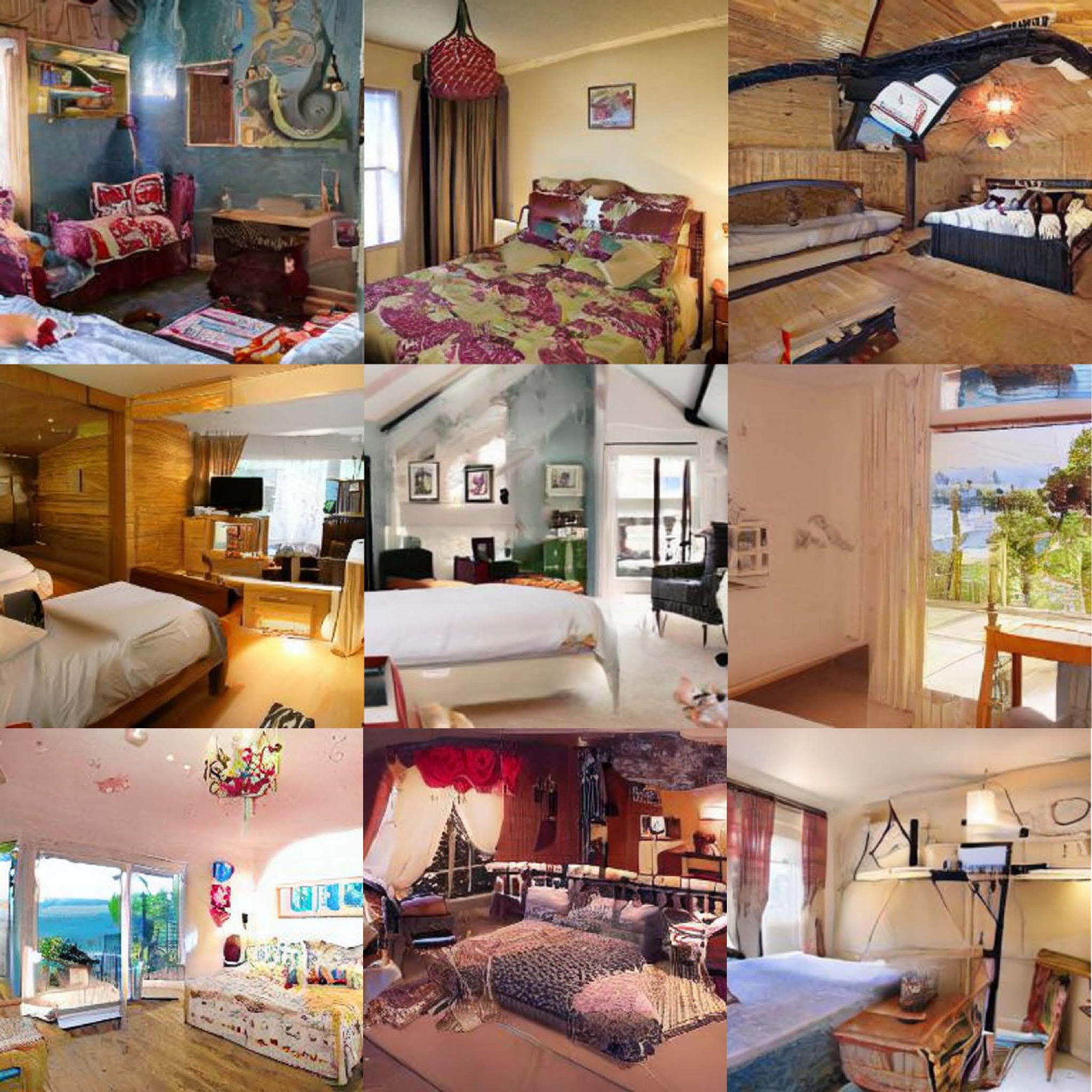}
    \caption{ADP-DM(6-bit)}
    \label{fig:4c}
  \end{subfigure}
  \caption{$256\times256$ LSUN-Bedroom samples from 100 step LDMs in 6-bit with different post-training quantization methods.}
  \label{Bedroom}
  \vspace{-0.3cm}
\end{figure*}

\begin{figure*}[t]
  \centering
  \begin{subfigure}{0.33\linewidth}
    \includegraphics[width=1\linewidth, height=1\textwidth]{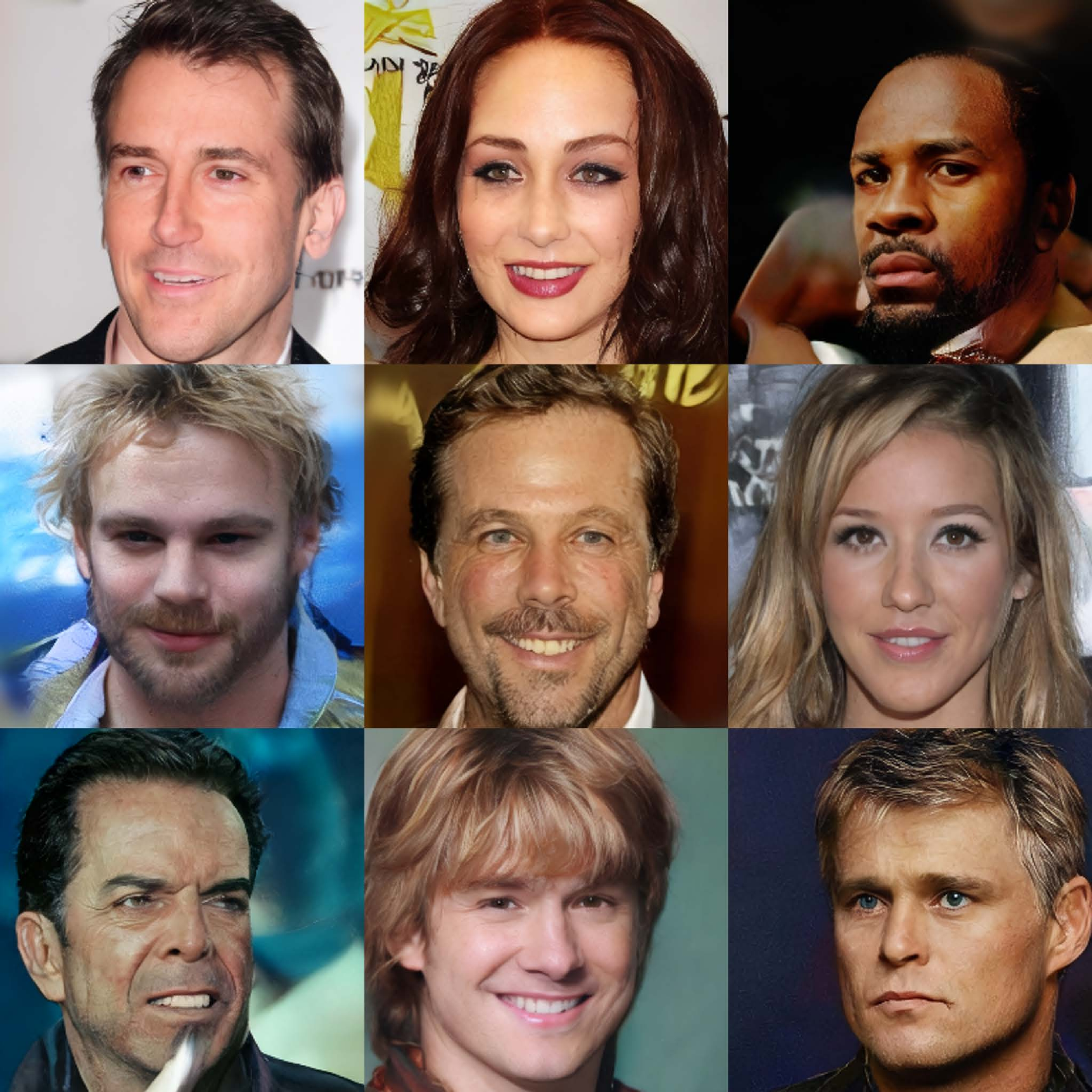}
    \caption{Full Precision}
    \label{fig:4a}
  \end{subfigure}
  \begin{subfigure}{0.33\linewidth}
    \includegraphics[width=1\linewidth, height=1\textwidth]{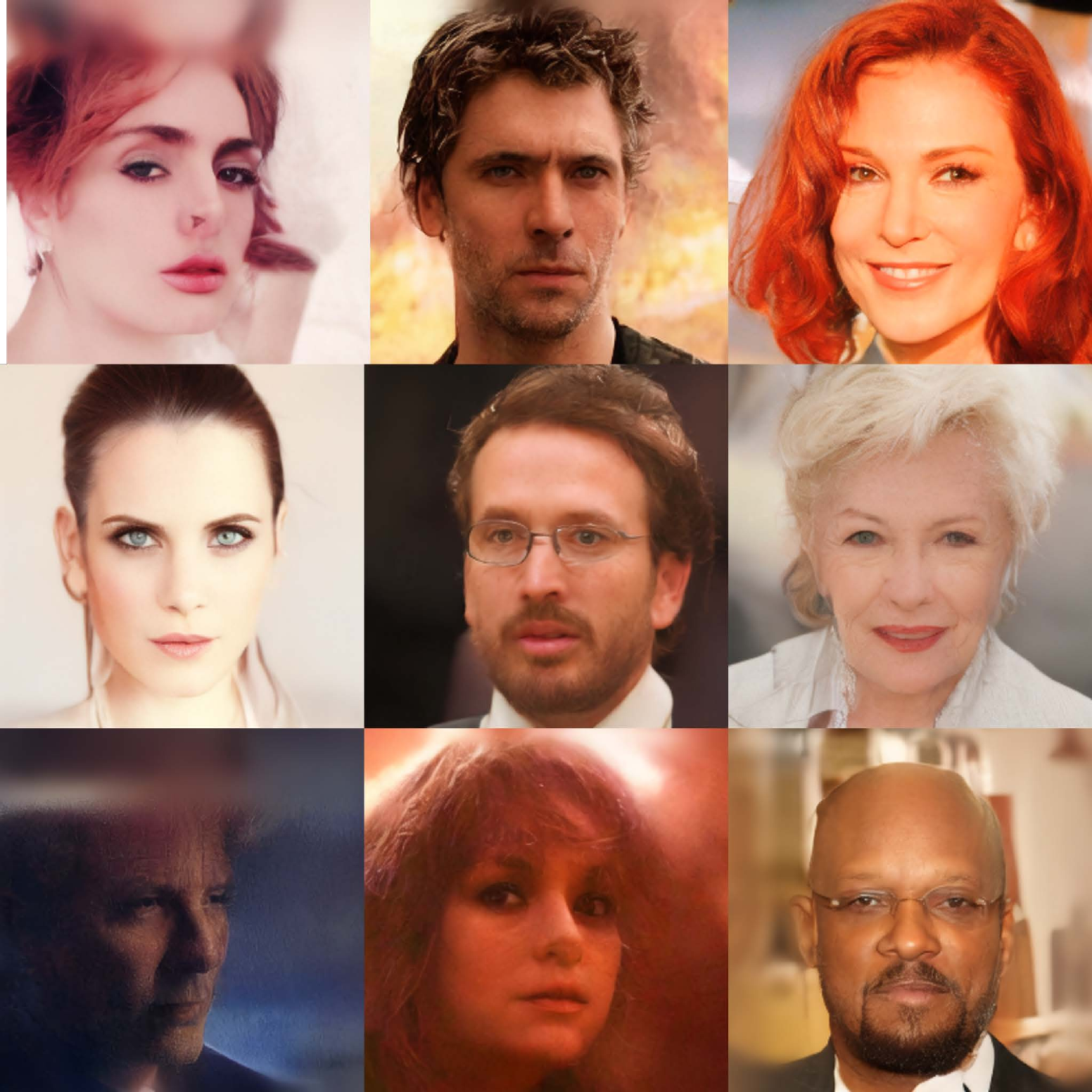}
    \caption{PTQ4DM(6-bit)}
    \label{fig:4b}
  \end{subfigure}
  \begin{subfigure}{0.33\linewidth}
    \includegraphics[width=1\linewidth, height=1\textwidth]{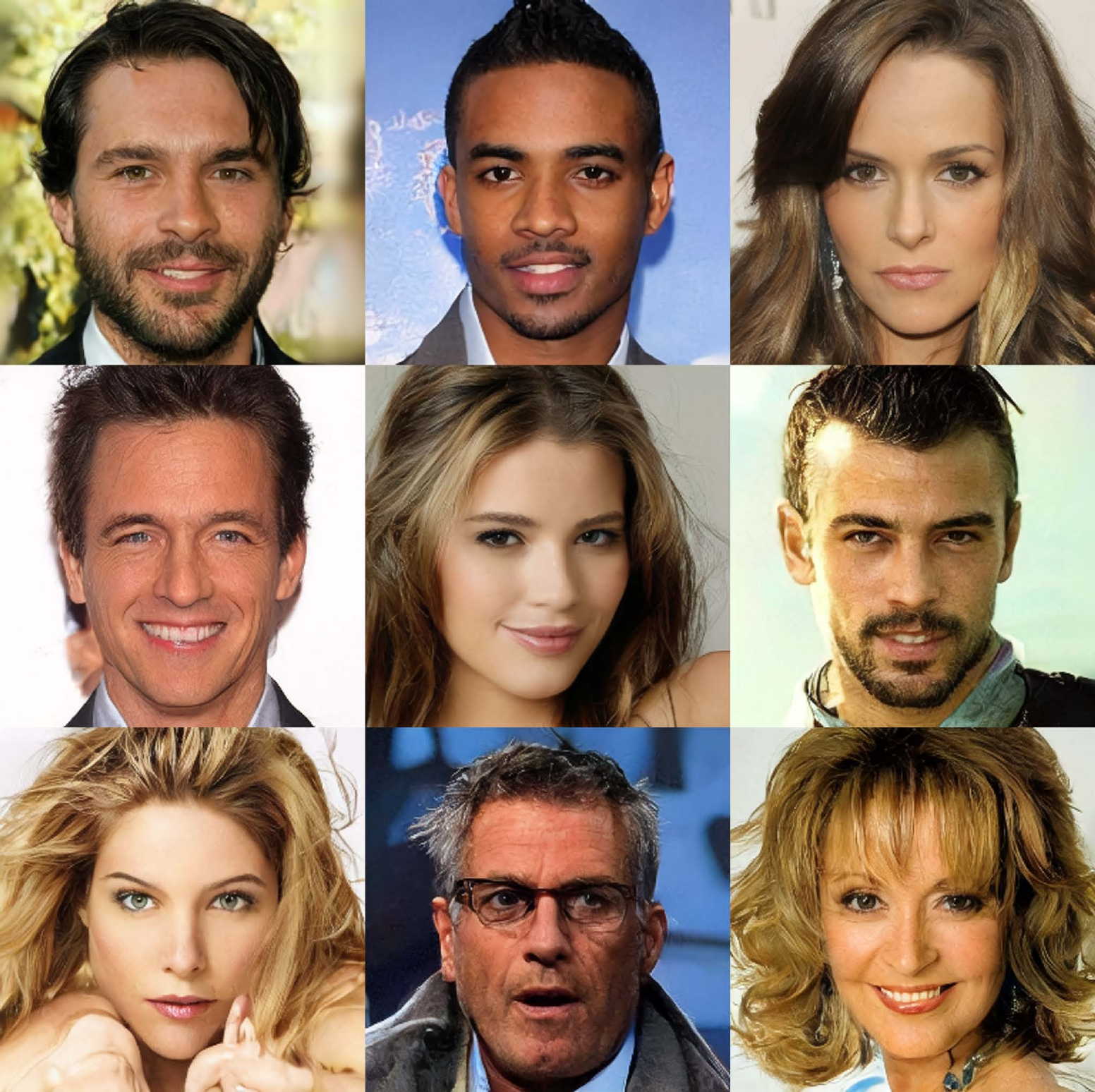}
    \caption{ADP-DM(6-bit)}
    \label{fig:4c}
  \end{subfigure}
  \caption{$256\times256$ CelebA-HQ samples from 100 step LDMs in 6-bit with different post-training quantization methods.}
  \label{CelebA-HQ}
  \vspace{-0.1cm}
\end{figure*}

\begin{figure*}[t]
	\centering
 \includegraphics[width=1\linewidth, height=1\textwidth]{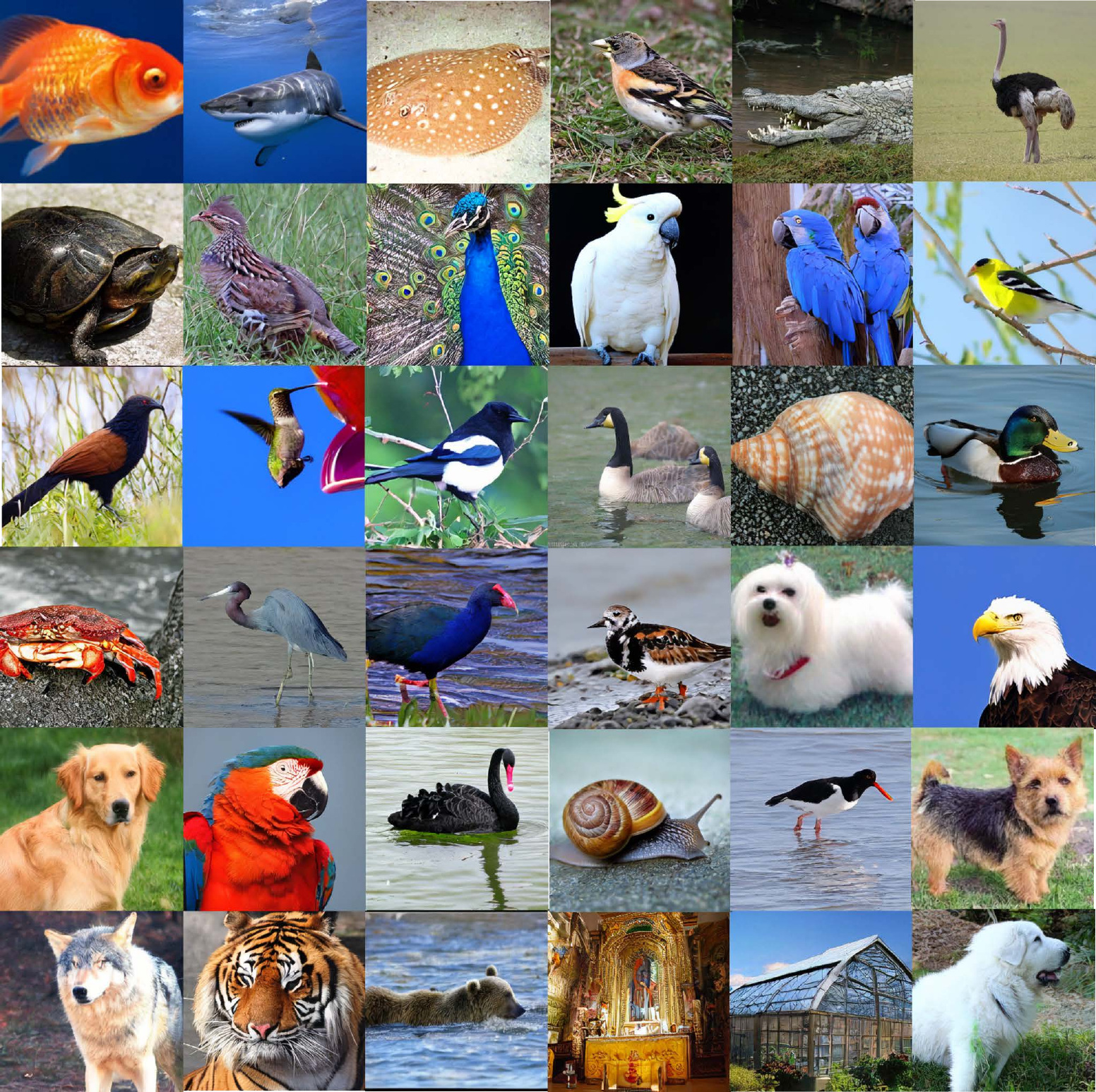}
	\caption{$256\times256$ ImageNet samples from 100 step LDMs in 6-bit with APQ-DM.}
	\vspace{-0.3cm}    
	\label{ImageNet}
\end{figure*}

\section{Samples}
\textbf{Additional samples: }We show more samples generated by the 6-bit quantized LDM-4 diffusion model with different post-training quantization methods in Figure \ref{Church} ($256\times256$ Church), Figure \ref{Bedroom} ($256\times256$ Bedroom), Figure \ref{CelebA-HQ} ($256\times256$ CelebA-HQ), and Figure \ref{ImageNet} ($256\times256$ ImageNet). Compared with the conventional quantization method in diffusion models, our APQ-DM can still achieve high-quality details in plausible images for various datasets with weights and activations in low bitwidths, which are semblable to the full precision ones.

\end{document}